\documentclass[lettersize,journal]{IEEEtran}
\usepackage{amsmath,amsfonts}
\usepackage{algorithmic}
\usepackage{algorithm}
\usepackage{array}
\usepackage[caption=false,font=footnotesize,labelfont=rm,textfont=rm]{subfig}
\usepackage{textcomp}
\usepackage{stfloats}
\usepackage{url}
\usepackage{verbatim}
\usepackage{graphicx}
\usepackage{cite}
\usepackage{nicefrac}
\usepackage{multirow}
\usepackage{booktabs}
\usepackage[breaklinks,colorlinks,linkcolor=blue,citecolor=blue,urlcolor=blue]{hyperref}
\usepackage{orcidlink}
\usepackage{simpleicons}
\usepackage{fontawesome5}

\newcommand{\tablestyle}[2]{\setlength{\tabcolsep}{#1}\renewcommand{\arraystretch}{#2}\centering\footnotesize}

\usepackage{color, colortbl}
\definecolor{Gray}{gray}{0.9}

\newcommand{\DIORRResults}{
\begin{table*}[!htbp]
    \setlength{\tabcolsep}{1.8pt}
    \caption{Comparison with state-of-the-art methods on the \textbf{DIOR-R}. The results in \textbf{bold} denote the best performance of each column.}
    \resizebox{\linewidth}{!}
    {
    \begin{tabular}{l|c|cccccccccccccccccccc|c}
    \toprule
    Mehtod & Backbone  & APL  & APO  & BF   & BC   & BR   & CH   & DAM  & ETS  & ESA  & GF   & GTF  & HA   & OP   & SH   & STA  & STO  & TC   & TS   & VE   & WM   & AP$_{50}$  \\
    \midrule
    \textit{one-stage:} &  &  &  &  &  &  &  &  &  &  &  &  &  &  &  &  &  &  &  &  &  &  \\
    RetinaNet-O~\cite{focal_loss} &R50   & 61.49    & 28.52    & 73.57    & 81.17    & 23.98    & 72.54    & 19.94    & 72.39    & 58.20    & 69.25    & 79.54    & 32.14    & 44.87    & 77.71    & 67.57    & 61.09    & 81.46    & 47.33    & 38.01    & 60.24    & 57.55    \\
    DFDet~\cite{dfdet}  & R50 & 61.92 & 38.83 & 77.41 & 81.36 & 34.11 & 74.97 & 26.26 & 62.31 & 76.06 & 75.56 & 79.62 & 38.26 & 52.76 & 80.40 & 73.11 & 68.27 & 81.38 & 52.23 & 44.11 & 63.35 & 62.11 \\
    Oriented Rep~\cite{oriented_reppoints} & R50  & \textbf{70.03} & 46.11    & 76.12    & 87.19    & 39.14    & \textbf{78.76}    & 34.57    & 71.80    & 80.42    & 76.16    & 79.41    & 45.48    & 54.90    & 87.82    & 77.03    & 68.07    & 81.60    & 56.83    & 51.57    & \textbf{71.25} & 66.71    \\
    DCFL~\cite{dcfl} & R50  & 68.60    & 53.10    & 76.70 & 87.10    & 42.10    & 78.60 & 34.50    & 71.50    & 80.80    & \textbf{79.70} & 79.50    & 47.30    & 57.40    & 85.20    & 64.60    & 66.40    & 81.50    & 58.90    & 50.90    & 70.90    & 66.80    \\
    \midrule
    \textit{two-stage:} &  &  &  &  &  &  &  &  &  &  &  &  &  &  &  &  &  &  &  &  &  &  \\
    Gliding Vertex~\cite{gliding_vertex} & R50   & 65.35    & 28.87    & 74.96    & 81.33    & 33.88    & 74.31    & 19.58    & 70.72    & 64.70    & 72.30    & 78.68    & 37.22    & 49.64    & 80.22    & 69.26    & 61.13    & 81.49    & 44.76    & 47.71    & 65.04    & 60.06    \\
    RoI Transformer~\cite{roi_transformer} & R50 & 63.34    & 37.88    & 71.78    & 87.53    & 40.68    & 72.60    & 26.86    & 78.71    & 68.09    & 68.96    & 82.74    & 47.71    & 55.61    & 81.21    & \textbf{78.23} & 70.26    & 81.61    & 54.86    & 43.27    & 65.52    & 63.87    \\
    QPDet~\cite{QPDet} & R50 & 63.22    & 41.39    & 71.97    & \textbf{88.55} & 41.23    & 72.63    & 28.82    & \textbf{78.90} & 69.00    & 70.07    & \textbf{83.01} & 47.83 & 55.54    & 81.23    & 72.15    & 62.66    & \textbf{89.05} & 58.09    & 43.38    & 65.36    & 64.20    \\
    AOPG~\cite{dior} & R50  & 62.39    & 37.79    & 71.62    & 87.63    & 40.90    & 72.47    & 31.08    & 65.42    & 77.99    & 73.20    & 81.94    & 42.32    & 54.45    & 81.17    & 72.69    & 71.31    & 81.49    & 60.04 & 52.38    & 69.99    & 64.41    \\
    DODet~\cite{dodet}      & R50 & 63.40 & 43.35 & 72.11 & 81.32 & \textbf{43.12} & 72.59 & 33.32 & 78.77 & 70.84 & 74.15 & 75.47 & \textbf{48.00} & \textbf{59.31} & 85.41 & 74.04 & 71.56 & 81.52 & 55.47 & 51.86 & 66.40 & 65.10 \\
    \midrule
    \textit{end-to-end:} & &  &  &  &  &  &  &  &  &  &  &  &  &  &  &  &  &  &  &  &  &  \\
    ARS-DETR~\cite{arsdetr} & R50  & 68.00    & 54.17    & 74.43    & 81.65    & 41.13    & 75.66    & 34.89    & 73.07    & 81.92 & 76.10 & 78.62  & 36.33    & 55.41    & 84.55    & 70.09    & 72.23    & 81.14    & 61.52    & 50.57    & 70.28    & 66.12    \\
    \midrule
    \textit{end-to-end:} & &  &  &  &  &  &  &  &  &  &  &  &  &  &  &  &  &  &  &  &  &  \\
    OrientedFormer & LSK-T        & 58.89 & 42.64 & 78.56 & 84.56 & 37.49 & 74.05 & 33.39 & 71.97 & 79.25 & 74.33 & 80.73 & 43.10 & 52.84 & 87.89 & 66.08 & 68.85 & 86.61 & 58.30 & 55.48 & 66.31 & 65.07 \\
    OrientedFormer & R50 & 65.65    & 48.69 & \textbf{78.79}    & 87.17    & 41.90 & 76.34    & 34.37 & 72.14    & 81.40    & 75.34    & 79.83    & 45.15    & 56.12 & \textbf{88.66} & 67.59    & 72.68 & 87.32    & 60.31    & 56.54 & 69.56    &67.28 \\
    OrientedFormer & Swin-T & 67.45 & \textbf{50.81} & 78.47 & 86.19 & 42.92 & 77.88 & \textbf{40.76} & 75.52 & \textbf{84.13} & 77.86 & 81.76 & 46.19 & 56.53 & 88.64 & 75.32 & \textbf{73.81} & 86.72 & \textbf{60.80} & \textbf{56.88} & 68.09 & \textbf{68.84} \\
    \bottomrule

    \end{tabular}
    }
    \label{DIOR-R-result}
    \end{table*}
}

\newcommand{\DOTAOneResults}{
    \begin{table*}[!htbp]
  \caption{Comparison with state-of-the-art methods on the \textbf{DOTA-v1.0} dataset. * denotes multi-scale trainig and testing. The results in \textbf{bold} denote the best performance of each column.}
  \large
  \resizebox{\linewidth}{!}
  {
  \begin{tabular}{l|c|ccccccccccccccc|c}
  \toprule
  Method   & Backbone & PL    & BD    & BR    & GTF   & SV    & LV    & SH    & TC    & BC    & ST    & SBF   & RA    & HA    & SP    & HC    & AP$_{50}$   \\
  \midrule
  \textit{one-stage:}  &    & & & & & & & & & & & & & & & & \\
  PSC~\cite{psc}      & R50      & 88.24 & 74.42 & 48.63 & 63.44 & 79.98 & 80.76 & 87.59 & 90.88 & 82.02 & 71.58 & 59.12 & 60.78 & 65.78 & 71.21 & 53.06 & 71.83 \\
  R3Det~\cite{r3det}    & R101     & 88.76 & 83.09 & 50.91 & 67.27 & 76.23 & 80.39 & 86.72 & 90.78 & 84.68 & 83.24 & 61.98 & 61.35 & 66.91 & 70.63 & 53.94 & 73.79 \\
  S$^{2}$A-Net~\cite{s2anet}      & R50 & 89.11 & 82.84 & 48.37 & 71.11 & 78.11 & 78.39 & 87.25 & 90.83 & 84.90 & 85.64 & 60.36 & 62.60 & 65.26 & 69.13 & 57.94 & 74.12 \\
  H2RBox~\cite{h2rbox} & R50     &88.93 & 78.89 & 46.27 & 68.79 & 81.12 & 75.45 & 86.68 & 90.89 & 86.71 & \textbf{87.33} & 64.15 & 68.83 & 62.81 & 69.39 & 59.79 & 74.40 \\
  CFA~\cite{cfa}      & R50      & 88.34 & 83.09 & 51.92 & 72.23 & 79.95 & 78.68 & 87.25 & 90.90 & 85.38 & 85.71 & 59.63 & 63.05 & 73.33 & 70.36 & 47.86 & 74.51 \\
  DHRec~\cite{dhrec}   & R50    & 88.58 & 77.90 & 53.84 & 72.93 & 78.45 & 78.84 & 87.64 & 90.88 & 88.78 & 85.46 & 56.11 & 66.74 & 67.58 & 70.25 & 57.53 & 74.57 \\
  DFDet~\cite{dfdet}   & R50    &88.92 & 79.25 & 48.40 & 70.00 & 80.22 & 78.85 & 87.21 & 90.90 & 83.13 & 83.98 & 60.07 & 66.49 & 68.27 & 76.78 & 58.11 & 74.71 \\
  SASM~\cite{sasm}     & R50      & 86.42 & 78.97 & 52.47 & 69.84 & 77.30 & 75.99 & 86.72 & 90.89 & 82.63 & 85.66 & 60.13 & 68.25 & 73.98 & 72.22 & 62.37 & 74.92 \\
  \midrule
  \textit{two-stage:}  &    & & & & & & & & & & & & & & & & \\
  RoI Transformer~\cite{roi_transformer}      & R50      & 88.65 & 82.60 & 52.53 & 70.87 & 77.93 & 76.67 & 86.87 & 90.71 & 83.83 & 52.81 & 53.95 & 67.61 & 74.67 & 68.75 & 61.03 & 74.61 \\
  Gliding Vertex~\cite{gliding_vertex} & R101     & 89.64 & \textbf{85.00} & 52.26 & 77.34 & 73.01 & 73.14 & 86.82 & 90.74 & 79.02 & 86.81 & 59.55 & \textbf{70.91} & 72.94 & 70.86 & 57.32 & 75.02 \\
  AOPG~\cite{dior} & R50 & 89.27 & 83.49 & 52.50 & 69.97 & 73.51 & 82.31 & 87.95 & 90.89 & 87.64 & 84.71 & 60.01 & 66.12 & 74.19 & 68.30 & 57.80 & 75.24 \\
  DODet~\cite{dodet}  & R50 & 89.34 & 84.31 & 51.39 & 71.04 & 79.04 & 82.86 & 88.15 & 90.90 & 86.88 & 84.91 & 62.69 & 67.63 & 75.47 & 72.22 & 45.54 & 75.49 \\
  Oriented R-CNN~\cite{orientedrcnn} & R50      & 89.46 & 82.12 & 54.78 & 70.86 & 78.93 & 83.00 & 88.20 & 90.90 & 87.50 & 84.68 & 63.97 & 67.69 & 74.94 & 68.84 & 52.28 & 75.87 \\
  SCRDet++ * ~\cite{SCRDetplusplus}         & R101 & \textbf{89.77} & 83.90 & 56.30 & 73.98 & 72.60 & 75.63 & 82.82 & 90.76 & \textbf{87.89} & 86.14 & 65.24 & 63.17 & 76.05 & 68.06 & 70.24 & 76.20 \\
  \midrule
  \textit{end-to-end:} &    & & & & & & & & & & & & & & & & \\
  D. DETR-O~\cite{deformabledetr}     & R50      & 84.89 & 70.71 & 46.04 & 61.92 & 73.99 & 78.83 & 87.71 & 90.07 & 77.97 & 78.41 & 47.07 & 54.48 & 66.87 & 67.66 & 55.62 & 69.48 \\
  D. DETR-O w/ CSL~\cite{csl}     & R50      & 86.27 & 76.66 & 46.64 & 65.29 & 76.80 & 76.32 & 87.74 & 90.77 & 79.38 & 82.36 & 54.00 & 61.47 & 66.05 & 70.46 & 61.97 & 72.15 \\
  ARS-DETR~\cite{arsdetr} & R50      & 86.97 & 75.56 & 48.32 & 69.20 & 77.92 & 77.94 & 87.69 & 90.50 & 77.31 & 82.86 & 60.28 & 64.58 & 74.88 & 71.76 & 66.62 & 74.16 \\
  \midrule
  \textit{end-to-end:} &    & & & & & & & & & & & & & & & & \\
  OrientedFormer& R50      & 88.14 & 79.13 & 51.96 & 67.34 & 81.02 & 83.26 & 88.29 & \textbf{90.90} & 85.57 & 86.25 & 60.84 & 66.36 & 73.81 & 71.23 & 56.49 & 75.37 \\
  OrientedFormer& Swin-T & 88.74 & 78.94 & 53.43 & 72.05 & 81.08 & 84.22 & 88.40 & 90.90 & 86.23 & 86.65 & 61.05 & 63.11 & 75.78 & 73.02 & 54.62 & 75.88 \\
  OrientedFormer& R101 & 88.18 & 82.14 & 52.60 & 72.00 & 80.88 & 83.11 & 88.35 & 90.87 & 84.08 & 86.31 & 63.18 & 67.26 & 76.58 & 69.08 & 54.12 & 75.92 \\
  OrientedFormer*    & R50      & 87.92 & 83.29 & \textbf{58.92} & \textbf{80.90} & \textbf{81.93} & \textbf{84.62} & \textbf{88.81} & 90.89 & 86.81 & 86.95 & \textbf{66.68} & 59.86 & \textbf{78.82} & \textbf{77.88} & \textbf{71.68} & \textbf{79.06} \\
  \bottomrule
  \end{tabular}
  }

  \label{DOTA-1.0-result}
    \end{table*}
}
\newcommand{\TableDOTAVonefiveResult}{
    \begin{table}[!t]
    \renewcommand\arraystretch{1.1}
    \caption{Main results of small size objects on \textbf{DOTA-v1.5}. The results in \textbf{bold} denote the best performance of each column.}
    \resizebox{\linewidth}{!}
    {

    \begin{tabular}{l|cccccc|c}
    \toprule
    Method        & SV    & LV    & SH    & ST    & SP    & CC    & AP$_{50}$   \\ \midrule
    RetinaNet-O~\cite{focal_loss}   & 44.53 & 56.79 & 73.31 & 59.96 & 64.52 & 0.83  & 59.16 \\
    Faster RCNN-O~\cite{faster_rcnn} & 51.28 & 68.98 & 79.37 & 67.50 & 65.28 & 1.54  & 62.00 \\
    Mask R-CNN~\cite{mask_rcnn}    & 51.31 & 71.34 & 79.75 & 66.07 & 64.46 & 9.42  & 62.67 \\
    HTC~\cite{htc}           & 51.54 & 73.31 & 80.31 & 67.34 & 64.48 & 5.15  & 63.40 \\
    ReDet~\cite{redet}         & 52.38 & 75.73 & 80.92 & 68.64 & 70.55 & \textbf{11.53} & 66.86 \\
    OrientedFormer    & \textbf{64.05} & \textbf{77.04} & \textbf{85.33} & \textbf{78.11} & \textbf{72.08} & 10.86 & \textbf{67.06} \\ \bottomrule
    \end{tabular}

    }

    \label{DOTA-1.5-result}
    \end{table}
}
\newcommand{\TableDOTAVtwoResult}{
    \begin{table}[!htbp]\large
    \renewcommand\arraystretch{1.2}
    \caption{Performance comparisons on the \textbf{DOTA-v2.0} dataset.}
    \large
    \resizebox{\linewidth}{!}
    {

    \begin{tabular}{c|cccc}
    \toprule
    Method & SASM~\cite{sasm}            & RetinaNet-O~\cite{focal_loss}    & Oriented Rep~\cite{oriented_reppoints}   & Mask R-CNN~\cite{mask_rcnn} \\
    AP$_{50}$    & 44.53           & 46.68          & 48.95          & 49.47      \\ \midrule
    Method & ATSS-O~\cite{atss}          & S$^{2}$A-Net~\cite{s2anet}        & HTC~\cite{htc}            & DCFL~\cite{dcfl}       \\
    AP$_{50}$    & 49.57           & 49.86          & 50.34          & 51.57      \\ \midrule
    Method & RoI Trans.~\cite{roi_transformer} & S$^{2}$A-Net$+$DCFL & Oriented R-CNN~\cite{orientedrcnn} & OrientedFormer \\
    AP$_{50}$    & 52.81           & 52.84          & 53.28          & \textbf{54.27}      \\ \bottomrule
    \end{tabular}

    }

    \label{DOTA-2.0-result}
    \end{table}
}
\newcommand{\HRSCResults}{
    \begin{table}[!htbp]
    \centering
    \caption{Comparison with the state-of-the-art methods on the \textbf{HRSC2016} dataset. 07 means evaluation under pascal voc2007 metric, and 12 means evaluation under pascal voc2012 metric.}
    \resizebox{0.9\linewidth}{!}{
        \begin{tabular}{c|c|cc}
        \toprule
        Methods  & Backbone & mAP(07) & mAP(12) \\
        \midrule
        PSC~\cite{psc}  & R50      & 85.65   &  -      \\
        RoI Transformer~\cite{roi_transformer} & R101     & 86.20    &  -      \\
        Gliding Vertex~\cite{gliding_vertex}  & R101     & 88.20    &  -      \\

        PIoU~\cite{piou}& DLA34    & 89.20   & -       \\
        CenterMap~\cite{centermap}       & R50      & -       & 92.8    \\
        R3Det~\cite{r3det}      & R101     & 89.26   & 96.01   \\
        CSL~\cite{csl}&  R101     & 89.62   & 96.10    \\
        H2RBox-v2~\cite{h2rboxv2}  & R50      & 89.66   &  -      \\
        GWD~\cite{gwd}      & R101     & 89.85   & 97.37   \\
        OSKDet~\cite{oskdet}   & R101     & 89.98   &  -      \\
        S$^{2}$A-Net~\cite{s2anet}  & R101     & 90.17   & 95.01   \\
        Oriented R-CNN~\cite{orientedrcnn} & R50      & 90.40   & 96.50    \\
        \midrule
        OrientedFormer  & R50      & 90.17   & 96.48   \\
        \bottomrule
        \end{tabular}
        }
    \label{hrsc2016_results}
    \end{table}
}

\newcommand{\PositionalEncoding}{
    \begin{table}[!htbp]
        \centering
        \caption{Comparsions of different \textbf{positional encoding} in the decoder.}

    \resizebox{\linewidth}{!}{
          \begin{tabular}{c|ccccc}
          \toprule
          PE  & -     & Deform. & DAB.   & Learnable & Gaussian   \\
          \midrule
          AP$_{50}$ & 66.85 & 65.97  & 66.03 & 64.27     & \textbf{67.28} \\
          \bottomrule
          \end{tabular}
    }

        \label{ablation_pe}
      \end{table}

}

\newcommand{\AttentionBias}{
    \begin{table}[!t]
        \centering
        \caption{Comparsions of different \textbf{self-attention} in the decoder.}

    \resizebox{\linewidth}{!}{
          \begin{tabular}{c|cccc}
          \toprule
          Self-Attention & -     & iof   & iou   & Wasserstein   \\
          \midrule
          AP$_{50}$ & 67.03 & 66.57 & 67.08 & \textbf{67.28} \\
          \bottomrule
          \end{tabular}
    }

            \label{ablation_self_attn}
      \end{table}
}

\newcommand{\Tabblelevels}{
    \begin{table}[!htbp]
        \centering
        \caption{Comparsions of \textbf{different layers} of Backbone on DIOR-R.}
        \large
        \resizebox{\linewidth}{!}{

            \begin{tabular}{c|c|c|ccccc}
            \toprule
            Method                                                                      & Backbone             & Layers & AP$_{50}$  & AP$_{75}$  & AP$_{50:95}$ & Params &FLOPs \\ \midrule
            \multirow{4}{*}{\begin{tabular}[c]{@{}c@{}}Oriented\\ -Former\end{tabular}} & \multirow{4}{*}{R50} & 1      & 56.21 & 33.54 & 33.23   & 41M & 287G \\
                                                                                        &                      & 2      & 60.69 & 39.86 & 38.35   & 41M & 297G \\
                                                                                        &                      & 3      & 66.37 & 44.14 & 42.35   & 42M & 315G \\
                                                                                        &                      & 4      & 67.28 & 44.13 & 42.66   & 44M & 325G \\ \bottomrule
            \end{tabular}
        }

            \label{ablation_levels}
      \end{table}
}

\newcommand{\EachProposedModuleOnDIOR}{
    \begin{table}[!htbp]
        \centering
        \caption{The effectiveness of proposed \textbf{individual modules} on DIOR-R.}
        \renewcommand\arraystretch{1.1}

    \resizebox{\linewidth}{!}{
        \begin{tabular}{c|ccc|ccc}
        \toprule
        \multirow{2}{*}{Methods}                                                   & \multirow{2}{*}{\begin{tabular}[c]{@{}c@{}}Gaussian\\ PE\end{tabular}} & \multirow{2}{*}{\begin{tabular}[c]{@{}c@{}}Wasserstein\\ Self-Attention\end{tabular}} & \multirow{2}{*}{\begin{tabular}[c]{@{}c@{}}Oriented\\ Cross-Attention\end{tabular}} & \multicolumn{3}{c}{DIOR-R} \\ \cmidrule{5-7}
                                                                                   &                                                                        &                                                                                       &                                                                                     & AP$_{50}$ & AP$_{75}$   & AP$_{50:95}$    \\ \midrule
        \multirow{6}{*}{\begin{tabular}[c]{@{}c@{}}Oriented\\ Former\end{tabular}} &                                                                        &                                                                                       &                                                                                     & 62.69     & 44.18 & 41.38 \\
                                                                                   & \checkmark                                                             &  \checkmark                                                                           &                                                                                     & 63.08     & 43.44 & 41.00 \\
                                                                                   &                                                                        &                                                                                       & \checkmark                                                                          & 65.78     & 43.69 & 41.87 \\
                                                                                   &                                                                        &  \checkmark                                                                           & \checkmark                                                                          & 66.85     & 46.22 & 43.73 \\
                                                                                   & \checkmark                                                             &                                                                                       & \checkmark                                                                          & 67.03     & 44.07 & 42.49 \\
                                                                                   & \checkmark                                                             &  \checkmark                                                                           & \checkmark                                                                          & 67.28     & 44.13 & 42.66 \\ \bottomrule
        \end{tabular}
    }

                \label{ablation_indivisual_on_DIOR}
      \end{table}
}

\newcommand{\EachProposedModuleOnDOTA}{
    \begin{table}[!t]
        \centering
        \caption{The effectiveness of \textbf{individual modules} on DOTA-v1.0.}
    \renewcommand\arraystretch{1.1}

    \resizebox{\linewidth}{!}{
        \begin{tabular}{c|ccc|ccc}
        \toprule
        \multirow{2}{*}{Methods}                                                   & \multirow{2}{*}{\begin{tabular}[c]{@{}c@{}}Gaussian\\ PE\end{tabular}} & \multirow{2}{*}{\begin{tabular}[c]{@{}c@{}}Wasserstein\\ Self-Attention\end{tabular}} & \multirow{2}{*}{\begin{tabular}[c]{@{}c@{}}Oriented\\ Cross-Attention\end{tabular}} & \multicolumn{3}{c}{DOTA-v1.0} \\ \cmidrule{5-7}
                                                                                   &                                                                        &                                                                                       &                                                                                     & AP$_{50}$   & AP$_{75}$   & AP$_{50:95}$     \\ \midrule
        \multirow{6}{*}{\begin{tabular}[c]{@{}c@{}}Oriented\\ Former\end{tabular}} &                                                                        &                                                                                       &                                                                                     & 73.81  & 47.74 & 45.40    \\
                                                                                   & \checkmark                                                             &  \checkmark                                                                           &                                                                                     & 74.55  & 49.26 & 46.28       \\
                                                                                   &                                                                        &                                                                                       & \checkmark                                                                          & 74.64  & 47.80 & 45.85       \\
                                                                                   &                                                                        &  \checkmark                                                                           & \checkmark                                                                          & 74.69  & 46.16 & 45.12  \\
                                                                                   & \checkmark                                                             &                                                                                       & \checkmark                                                                          & 74.76  & 48.95 & 45.97  \\
                                                                                   & \checkmark                                                             &  \checkmark                                                                           & \checkmark                                                                          & 75.37  & 46.39 & 45.01   \\ \bottomrule
        \end{tabular}
    }

                \label{ablation_indivisual_on_DOTA}
      \end{table}
}

\newcommand{\TableFPS}{
    \begin{table}[t]
    \centering
    \renewcommand\arraystretch{1.2}
    \caption{\textbf{Speed, Parameters, FLOPs and accuracy} on DOTA-v1.0.}
    \large
    \resizebox{\linewidth}{!}
    {
        \begin{tabular}{l|c|c|cccc}
        \toprule
        Method           & Frame & Backbone & FPS & Params & FLOPs & AP$_{50}$ \\ \midrule
        RoI Transformer~\cite{roi_transformer}& \multirow{3}{*}{\begin{tabular}[c]{@{}c@{}}Two-\\ stage\end{tabular}} &R50  &9.2 & 55M & 253G & 74.61 \\
        Oriented RCNN~\cite{orientedrcnn}     & &R50   &7.3 & 41M & 225G  & 75.87 \\
        Gliding Vertex~\cite{gliding_vertex}  & &R101  &10.2& 41M & 225G  & 75.02 \\ \midrule
        R3Det~\cite{r3det}           & \multirow{5}{*}{\begin{tabular}[c]{@{}c@{}}One-\\ stage\end{tabular}} &R101 &6.1 & 42M & 335G & 73.79\\
        CFA~\cite{cfa}               & &R50  &16.6& 37M  & 194G & 74.51 \\
        SASM~\cite{sasm}             & &R50  &15.8& 37M  & 194G & 74.92 \\
        PSC~\cite{psc}               & &R50  &16.8& 37M  & 218G & 71.83 \\
        DCFL~\cite{dcfl}             & &R50  &17.7& 36M  & 216G & 74.26 \\ \midrule
        D.DETR-O~\cite{deformabledetr}  & \multirow{3}{*}{\begin{tabular}[c]{@{}c@{}}End-\\ to-\\ end\end{tabular}}&R50  &10.8& 41M  &186G & 69.48\\
        D.DETR-O w/CSL~\cite{csl}       & &R50 &10.7& 41M  &186G & 72.15\\
        ARS-DETR~\cite{arsdetr}         & &R50 &10.1& 42M  &186G & 74.16\\ \midrule
        OrientedFormer & \multirow{3}{*}{\begin{tabular}[c]{@{}c@{}}End-\\ to-end\end{tabular}}&R50 &11.4& 44M &529G & 75.37\\
        OrientedFormer       &  &Swin-T  &11.2& 45M  &536G & 75.88\\
        OrientedFormer       &  &R101 &11.1 &63M &606G &75.92\\\bottomrule
        \end{tabular}
    }
    \label{FPS_Params}
    \end{table}
}

\newcommand{\TableExpSetting}{
    \begin{table}[t]
    \centering
    \caption{Experiment settings of OrientedFormer.}
            \begin{tabular}{c|l|l}
            \toprule
            Method                           & config              & value       \\ \midrule
            \multirow{14}{*}{OrientedFormer} & optimizer           & AdamW       \\
                                             & base learning rate  & 5e-5        \\
                                             & weight decay        & 1e-6        \\
                                             & optimizer momentum  & $\beta_{1},\beta_{2}$=0.9, 0.999 \\
                                             & batch size          & 4           \\
                                             & GPUs                & 2           \\
                                             & epochs              & 12 or 24    \\
                                             & lr decay epochs     & (8, 11) or (16, 22)     \\
                                             & warmup iter         & 500         \\
                                             & warmup factor       & 0.333       \\
                                             & clip gradient type  & full model  \\
                                             & clip gradient value & 1.0         \\
                                             & clip gradiant norm  & 2.0         \\
                                             & data augmentation   & RandomFlip  \\
                                             & seed                & Random Seed \\\bottomrule
            \end{tabular}
            \label{expsetting}
    \end{table}
}

\newcommand{\TableSamplingOffsets}{
    \begin{table}[!t]
    \centering
    \caption{Comparison of \textbf{Different Sampling} Methods on DIOR-R.}
            \begin{tabular}{c|c|cc}
            \toprule
                                            & Methods                  & AP$_{50}$  & AP$_{75}$  \\ \midrule
            \multirow{4}{*}{OrientedFormer} & Fixed Offsets            & 65.98 & 43.02 \\
                                            & Deformable Offsets       & 66.66 & 44.10 \\
                                            & Random Offsets           & 66.70 & 43.47 \\
                                            & Oriented Cross-attention & \textbf{67.28} & \textbf{44.13} \\ \bottomrule
            \end{tabular}
    \label{SamplingOffsets}
    \end{table}
}

\newcommand{\TableResultICDAR}{
    \begin{table}[!t]
    \centering
    \renewcommand\arraystretch{1.1}
    \caption{Main results of precision, recall (IoU=0.5), F-measure and Flops on \textbf{ICDAR2015}. The results in \textbf{bold} denote the best performance. The backbone used by all methods is Resnet50.}
        \begin{tabular}{l|cccc}
        \toprule
        Method        & Precision    & Recall    & F-measure & FLOPs  \\ \midrule
        SASM ~\cite{sasm} & 56.7 & 77.9 & 65.7 & 71G\\
        PSC ~\cite{psc} & 83.7 & 63.2 & 72.0   & 78G  \\
        Retinanet-O ~\cite{focal_loss} & 83.9 & 67.5 & 74.8 & 77G\\
        GWD ~\cite{gwd}& 84.4 & 67.6 & 75.1  &  77G     \\
        R3DET ~\cite{r3det} & 83.2 & 69.2 & 75.6  &  120G   \\
        Oriented RCNN~\cite{orientedrcnn}& 73.9 & \textbf{80.9} & 77.2 &  100G  \\
        CFA~\cite{cfa} & 78.3 & 77.1 & 77.7 & 71G   \\
        Oriented Reppoints~\cite{oriented_reppoints}& 79.0 & 77.0 & 78.0 &  71G    \\
        S$^{2}$A-Net~\cite{s2anet}& 81.6 & 75.6 & 78.5 &   77G    \\
        RoI Transformer ~\cite{roi_transformer} & 78.2 & 80.7 & 79.4  & 55G\\
         \midrule
        OrientedFormer    & \textbf{85.3} & 74.2 & \textbf{79.4} & 195G  \\ \bottomrule
        \end{tabular}
    \label{ResultICDAR2015}
    \end{table}
}

\newcommand{\notations}{
    \begin{table}[!htbp]
    \centering
    \renewcommand\arraystretch{1.1}
    \caption{Nomenclature with related notations.}
    \resizebox{\linewidth}{!}
    {
        \begin{tabular}{l|l|l|l}
        \toprule
        Notation & Description & Notation & Description \\
        \midrule
        $(x, y, w, h, \theta)$          & oriented boxes         &$d_{q}$                 & dimension of self-attention\\
        $(x, y, z, r, \theta)$          & oriented boxes           & $G$                    & Wasserstein distance score\\
        $\left\{f^l\right\}^L_{l=1}$    & levels of feature map      & $\tau$, $\epsilon$              & coefficient of $G$\\
        $H_{l}$                         & height of feature map      & $(\Delta x, \Delta y, \Delta z)$     & offset of sampling points   \\
        $W_{l}$                         & width of feature map       & $g$                                  & number of heads\\
        $D$                             & channel of feature map    &$O$                              & number of sampling points \\
        $N$                             & number of queries      & $(\tilde{x}, \tilde{y}, \tilde{z})$  & sampling points\\
        $Q_{c}$                         & content query            & $P$                                  & rotated sampling points \\
        $Q_{p}$                         & positional query         & $s^{l}$                       & downsampling stride\\
        $\varphi(\cdot )$               & positional encoding     & $\pi_{L}$                            & scale-aware attention\\
        $\mathbf{x}$                    & horizontal boxes   & $\pi_{C}$                & channel-aware attention\\
        $T$                             & temperature of PE      & $\pi_{S}$        & spatial-aware attention \\
        $K, k$                          & hyperparameter of PE      & $V$                    & value of cross-attnetion  \\
        $D'$                            & dimension of PE      & $V_{\pi_{L}}$                        & output of $\pi_{L}$\\
        $\mathcal{N}(\cdot, \cdot)$     & gaussian distribution   &  $V_{\pi_{C}}$                        & output of $\pi_{C}$   \\
        $\mu$                           & expectation    & $\eta$                  & coefficient of $\pi_{L}$ \\
        $\mathbf{\Sigma}$               & variance                   & $\gamma$                             & Linear \\
        $\mathbf{R}$                    & rotation matrix              &  $\rho$                               & ReLU      \\
        $\mathbf{\Lambda}$              & diagonal matrix              & $\mathcal{L} $ & loss function \\
          \bottomrule
        \end{tabular}
    }
    \label{notations}
    \end{table}
}

\newcommand{\QueriesNumber}{
    \newcolumntype{a}{>{\columncolor{Gray}}c}
    \resizebox{1.05\linewidth}{!}{
  \begin{tabular}{c|ccacc}
  \toprule
  Queries & 100   & 200   & \textbf{300}   & 400   & 500   \\
  \midrule
  AP$_{50}$     & 65.16 & 66.42 & \textbf{67.28} & 67.13 & 66.93 \\
  \bottomrule
  \end{tabular}
    }
}

\newcommand{\PointsNumber}{
    \newcolumntype{a}{>{\columncolor{Gray}}c}
    \resizebox{1.05\linewidth}{!}{
  \begin{tabular}{c|ccacc}
  \toprule
  Points      & 8     & 16    & \textbf{32}    & 64  & 128  \\
  \midrule
  AP$_{50}$    &65.60  &  67.02  & \textbf{67.28} & 66.27  & 66.97 \\
  \bottomrule
  \end{tabular}
    }
}

\newcommand{\HeadsNumber}{
    \newcolumntype{a}{>{\columncolor{Gray}}c}
    \resizebox{1.05\linewidth}{!}{
  \begin{tabular}{c|cccac}
  \toprule
  Heads & 8    & 16     & 32     & \textbf{64}    & 128    \\
  \midrule
  AP$_{50}$   & 66.33 & 66.63 & 66.73 & \textbf{67.28} & 66.78 \\
  \bottomrule
  \end{tabular}
    }
}

\newcommand{\TableAblationStudiesIntegrated}{
    \begin{table*}[!t]
  \centering
    \caption{\textbf{OrientedFormer ablation experiments} with ResNet-50 on DIOR-R. Default choice for our model is colored \colorbox{Gray}{gray}
  }

  \subfloat[\small
  The ablation of different number of \textbf{object queries}.
  \label{ablation_object_queries}
  ]{
      \begin{minipage}{0.3\linewidth}{
      \tablestyle{1.8pt}{1.2}
      \QueriesNumber
      }
      \end{minipage}

  }
        \hspace{1em}
    \subfloat[\small
  The ablation of different number of \textbf{sampling points} per head in oriented cross-attention.
  \label{ablation_sampling_points}
  ]{
      \begin{minipage}{0.3\linewidth}{
      \tablestyle{1.8pt}{1.2}
      \PointsNumber
      }
      \end{minipage}
  }
    \hspace{1em}
  \subfloat[\small
  The ablation of different number of \textbf{heads} in oriented cross-attention.
  \label{ablation_heads}
  ]{
      \begin{minipage}{0.3\linewidth}{
      \tablestyle{1.8pt}{1.2}
      \HeadsNumber
      }
      \end{minipage}
  }
  \label{ablations}
  \end{table*}
} 

\renewcommand{\algorithmicrequire}{\textbf{Input:} Images; Real Boxes; Real Classes.}
\renewcommand{\algorithmicensure}{\textbf{Output:} Loss $\mathcal{L}$.}
\newcommand{\Algorithmtrain}{
    \begin{algorithm}[!t]
        \footnotesize
        \caption{Training procedure}
        \label{Training_prodedure}
        \algorithmicrequire \\
        \algorithmicensure
        \begin{algorithmic}[1]
        \STATE  Decoder$\_$Layer $\leftarrow2$; Feature Layer $L\leftarrow5$; $\lambda_{cls}\leftarrow2$; $\lambda_{L_1}\leftarrow2$; $\lambda_{iou}\leftarrow 5$
        \STATE $\{f^l\}^L_{l=1}=$ Backbone$\_$with$\_$FPN(Images)
        \STATE $Q_c, Q_p$ = Initialize$\_$Query($\{f^l\}^L_{l=1}$)
        \FOR {$i = 0$ to Decoder$\_$Layer}
        \STATE $\varphi=$ Gaussian PE($Q_p$)
        \STATE $G=$ Gaussian Wasserstein distance score($Q_p$)
        \STATE $Q_c=$ Wasserstein Self-attention($Q_c, \varphi, G$)
        \STATE $Q_c=$ FFN(Oriented Cross-attention($Q_c, Q_p, \{f^l\}^L_{l=1}$))
        \STATE $Q_p=Q_p +$ {\tt{Linear1}} $(Q_c)$
        \STATE class$\_$score = {\tt{Linear2}} $(Q_c)$
        \STATE $Q_p^i$, class$\_$score$^i$ = Hungarian assignment( \{class$\_$score, Real Classes\},
                                                        \{$Q_p$, Real Boxes\})
        \STATE $\mathcal{L}_{cls}^i,\mathcal{L}_{1}^i,\mathcal{L}_{iou}^i=$ Focal loss(class$\_$score$^i$, Real Classes), $L_1$ loss($Q_p^i$, Real Boxes), Rotate IoU loss($Q_p^i$, Real Boxes)
        \ENDFOR
        \RETURN $\mathcal{L}= \sum \lambda_{cls}\mathcal{L}_{cls}^i+\lambda_{L_1}\mathcal{L}_{1}^i+\lambda_{iou}\mathcal{L}_{iou}^i$
        \end{algorithmic}
    \end{algorithm}

}

\usepackage{graphicx}

\newcommand{\overallarchitecture}{
    \begin{figure*}[t]
      \centering
       \includegraphics[width=\textwidth]{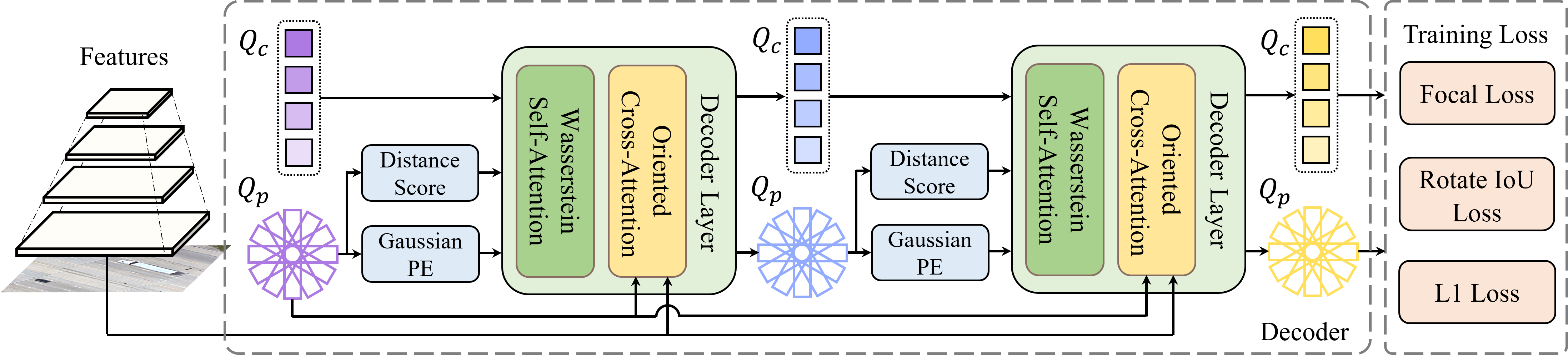}
       \caption{ Overall architecture of the OrientedFormer. Features are extracted from images. An object query is decomposed into a content query $Q_{c}$ and a positional query $Q_{p}$. The Gaussian PE encodes positional queries. The Wasserstein self-attention measures the geometric relations between two different content queries by utilizing Wasserstein distance scores. The oriented cross-attention is proposed to align values and positional queries.}
       \label{fig_overall_framework}
    \end{figure*}
}

\newcommand{\orientedthreedattn}{
    \begin{figure*}[t]
      \centering
       \includegraphics[width=\textwidth]{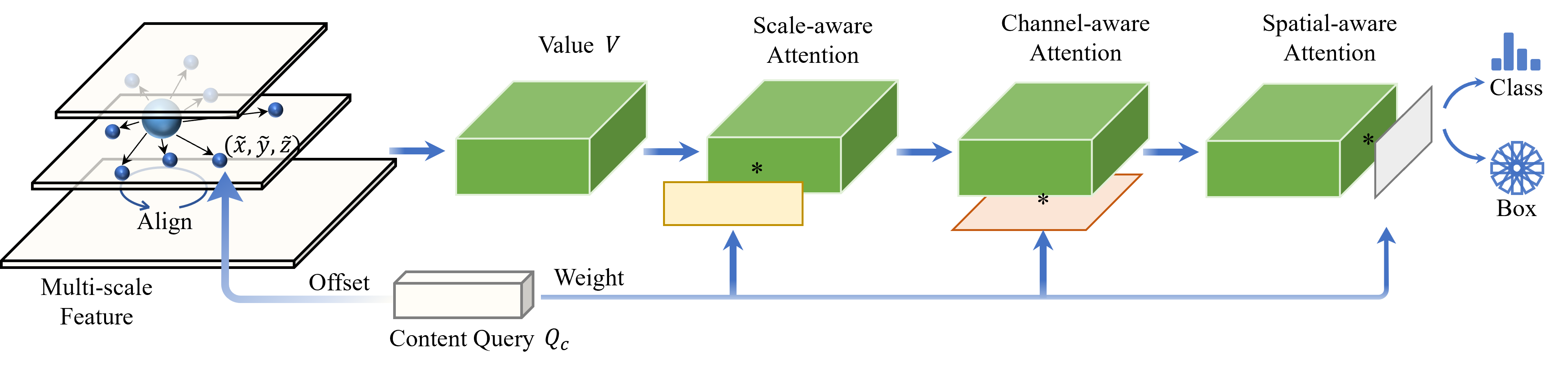}
       \caption{Oriented cross-attention. It attends to sparse sampling points $(\tilde{x},\tilde{y},\tilde{z})$ around the center of a positional query. Sampling points are rotated according to angles for alignment. Values $V$ are interpolated by sampling points and multi-scale features. We deploy attention mechanisms separately on each particular dimension of values, i.e., scale-aware, channel-aware, and spatial-aware.}
       \label{fig_oriented_3d_attn}
    \end{figure*}
}

\newcommand{\gaussianlikeselfattn}{
    \begin{figure}[t]
      \centering
      \subfloat[vanilla self-attention]
        {
            \includegraphics[width=0.45\linewidth]{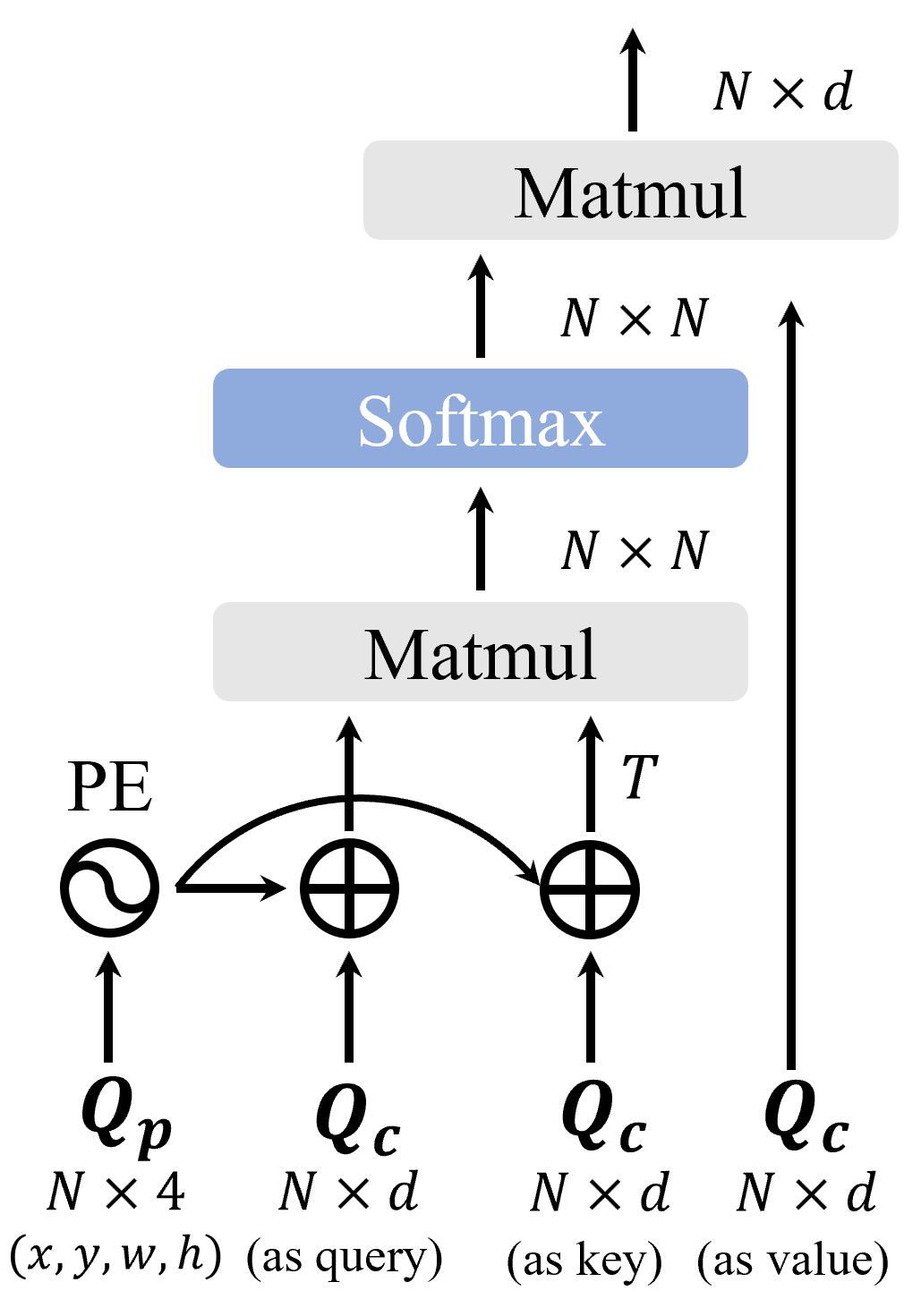}
            \label{vanilla_self_attn}
        }
      \subfloat[Wasserstein self-attention (ours)]
        {
            \includegraphics[width=0.455\linewidth]{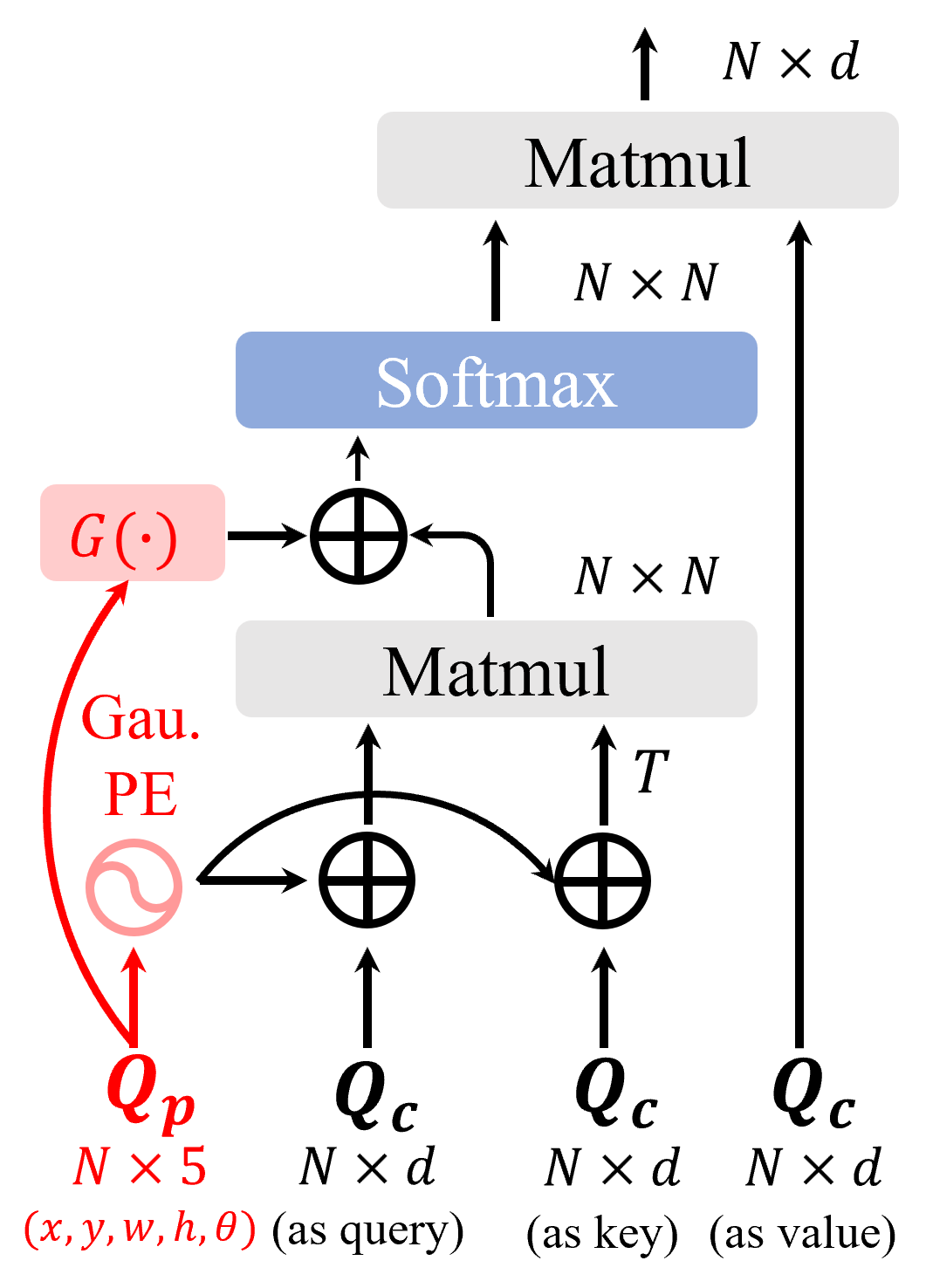}
            \label{wassersterin_self_attn}
        }
       \caption{Self-attention in the decoder. (a) vanilla self-attention. (b)Wasserstein self-attention.}
       \label{fig_gaussian_like_self-attn}
    \end{figure}
}

\newcommand{\gaussianlikepe}{
    \begin{figure}[t]
      \centering
            \subfloat[vanilla positional encoding]
        {
            \includegraphics[width=0.78\linewidth]{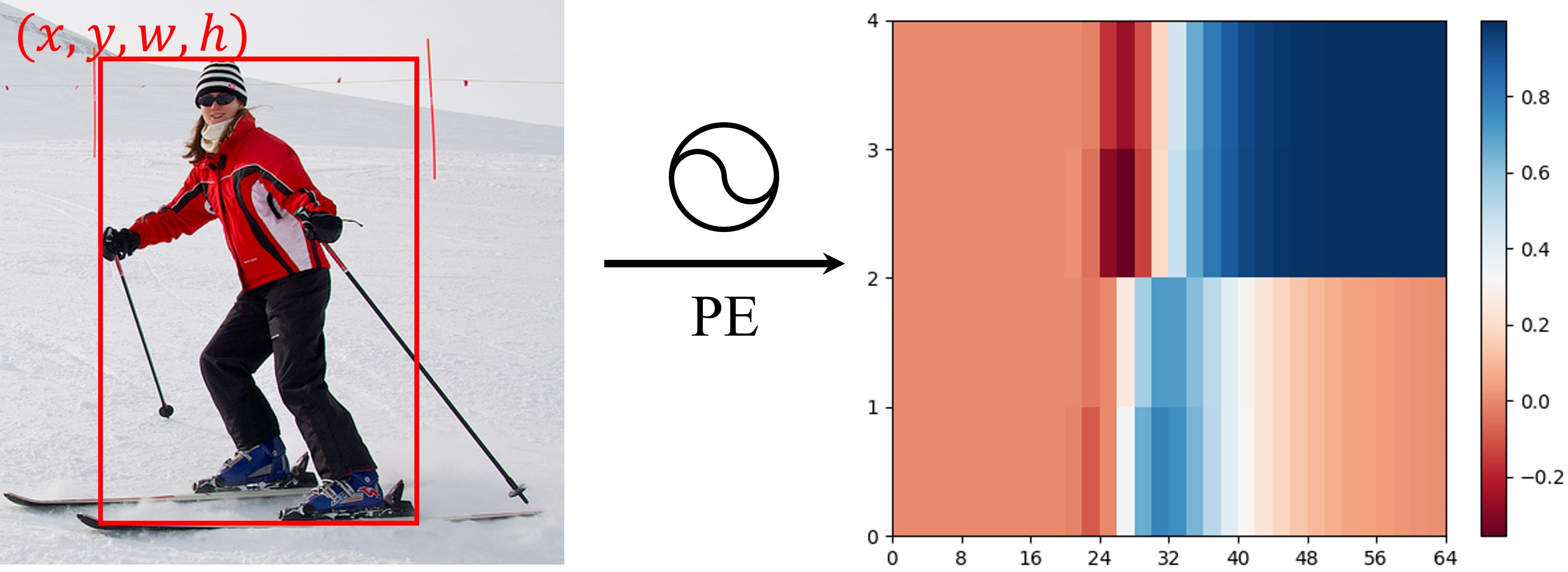}
            \label{vanilla_self_attn}
        }\\
      \subfloat[Gaussian positional encoding (ours)]
        {
            \includegraphics[width=0.99\linewidth]{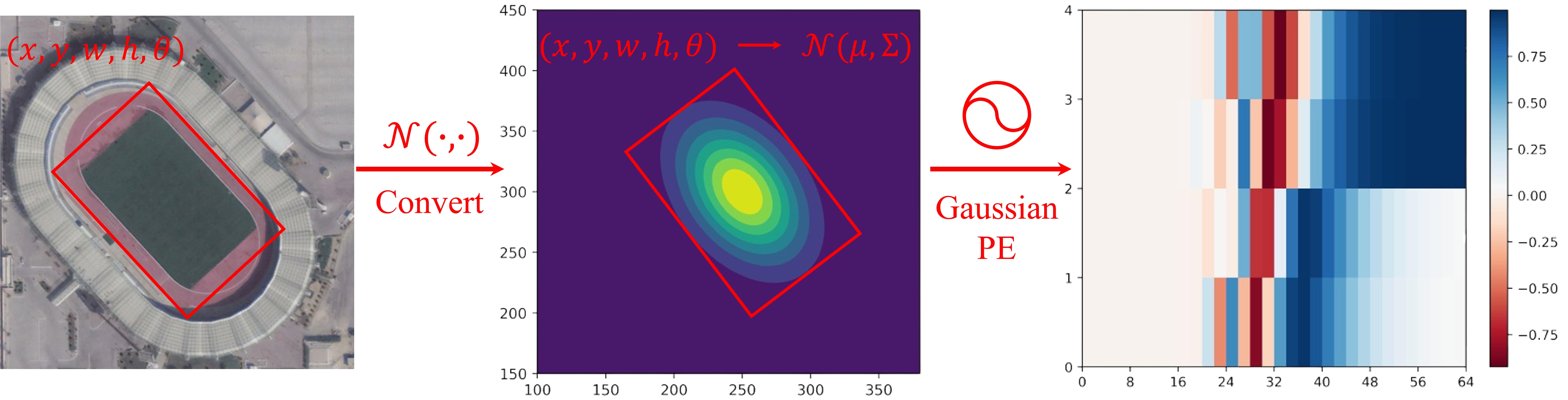}
            \label{wassersterin_self_attn}
        }
       \caption{An example of Gaussian positional encoding. (a) positional encoding in Deformable DETR. (b) Gaussian positional encoding.}
       \label{fig_gaussian_like_pe}
    \end{figure}
}

\newcommand{\visualresults}{
    \begin{figure*}[t]
      \centering
       \includegraphics[width=0.98\textwidth]{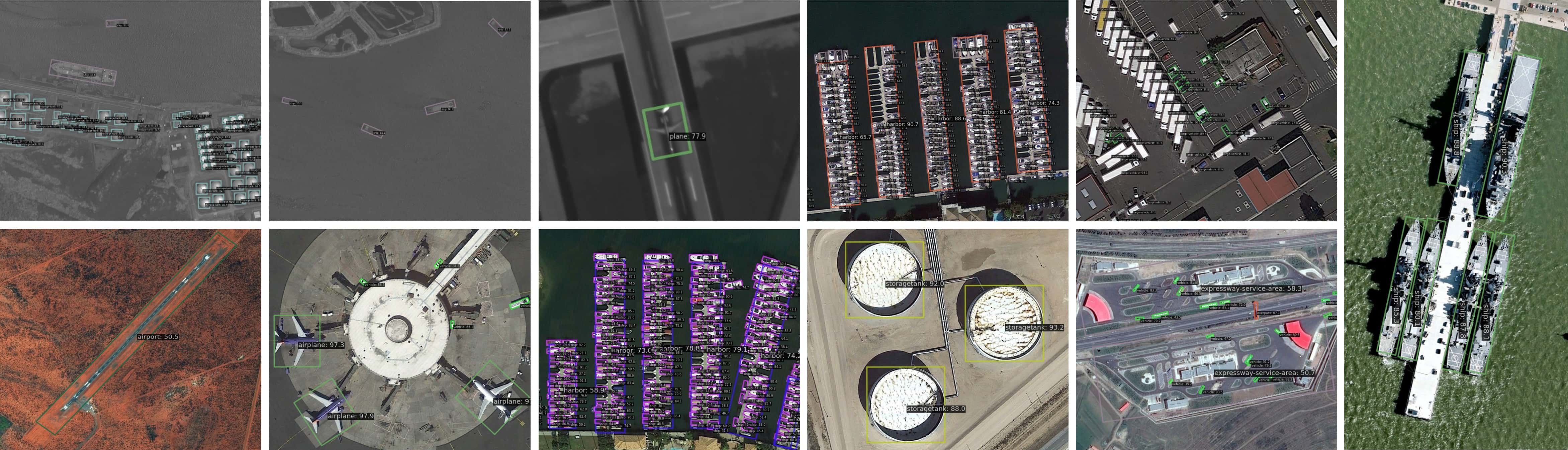}
       \caption{Visualization results of our method for DOTA, DIOR, and HRSC2016. DOTA contains images depicting extreme weather and poor lighting conditions. Oriented boxes, labels, and confidences are drawn.}
       \label{fig_visual_results}
    \end{figure*}
}

\newcommand{\visualsamplingpoint}{
    \begin{figure*}[t]
      \centering
       \includegraphics[width=0.98\textwidth]{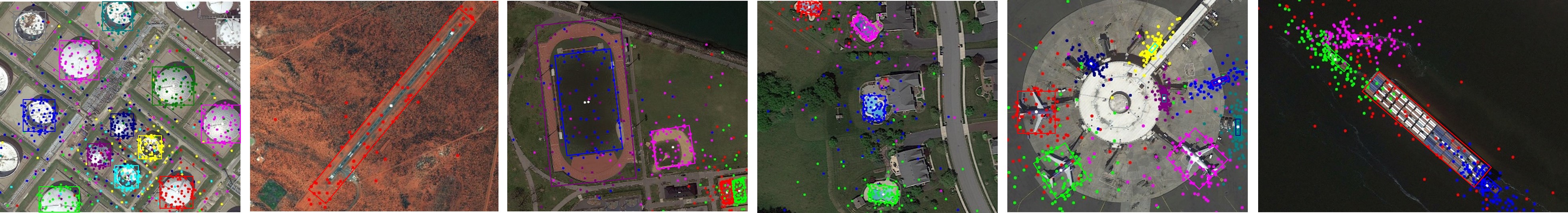}
       \caption{Visualization of oriented cross-attention. For readibility, we draw the center of positional queries and sampling points.}
       \label{fig_visual_sampling_point}
    \end{figure*}
}

\newcommand{\figureone}{
    \begin{figure}[!t]
    \centering
    \subfloat[\small]
        {
            \includegraphics[width=0.47\linewidth]{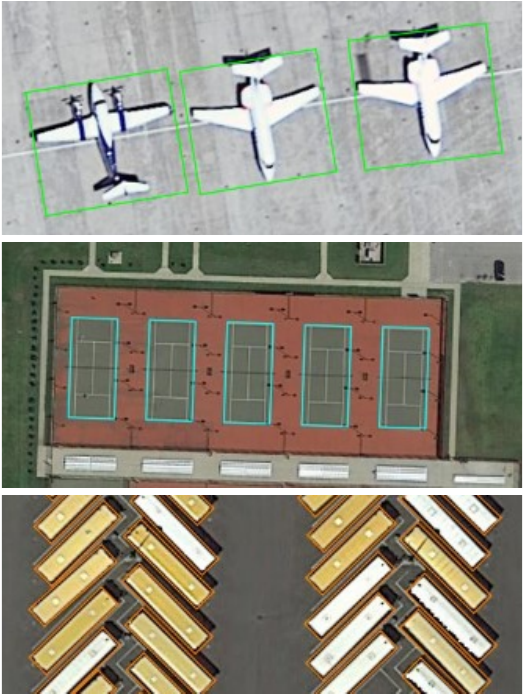}
            \label{fig_first_case}
        }
    \subfloat[\small]
        {
            \includegraphics[width=0.464\linewidth]{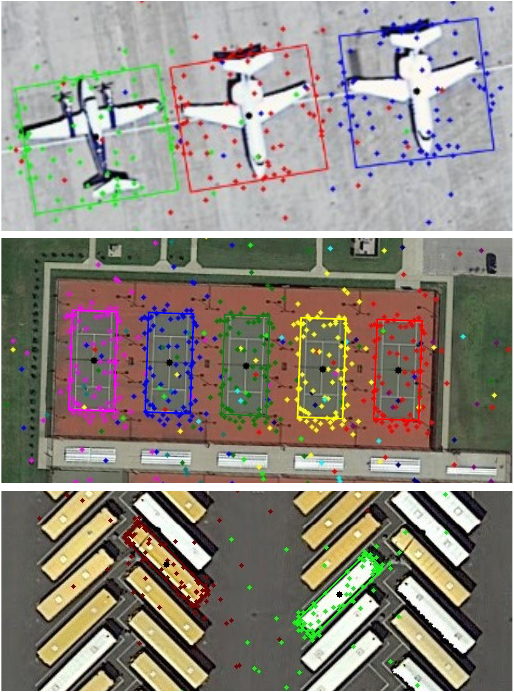}
            \label{fig_second_case}
        }
    \caption{(a) Object instances distribute in remote sensing images with arbitrary orientation. Angles are used to characterize oriented objects, in addition to positionsa and sizes. (b) Visualization of sampling points of the oriented cross-attention for alignment.}
    \label{fig_heatmap}
    \end{figure}

}

\newcommand{\convergence}{
    \begin{figure}[!t]
      \centering
       \includegraphics[width=1.0\linewidth]{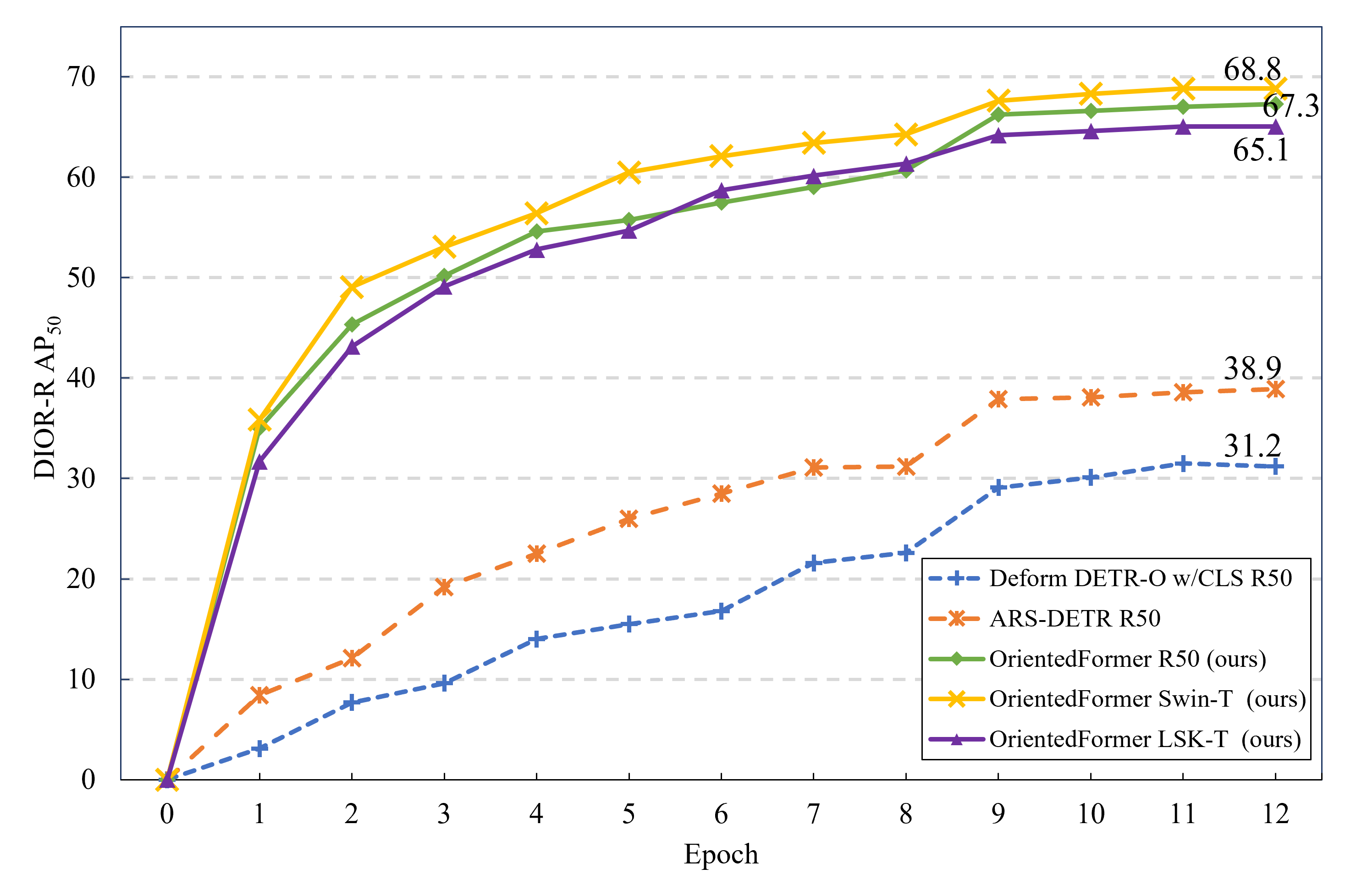}
       \caption{\textbf{Convergence curves} of Deformable DETR-O with CLS, ARS-DETR, OrientedFormer with ResNet50, Swin-T and LSK-T on DIOR-R.}
       \label{convergence_curve}
    \end{figure}
}

\newcommand{\compareepochsmap}{
    \begin{figure}[!t]
      \centering
       \includegraphics[width=1.0\linewidth]{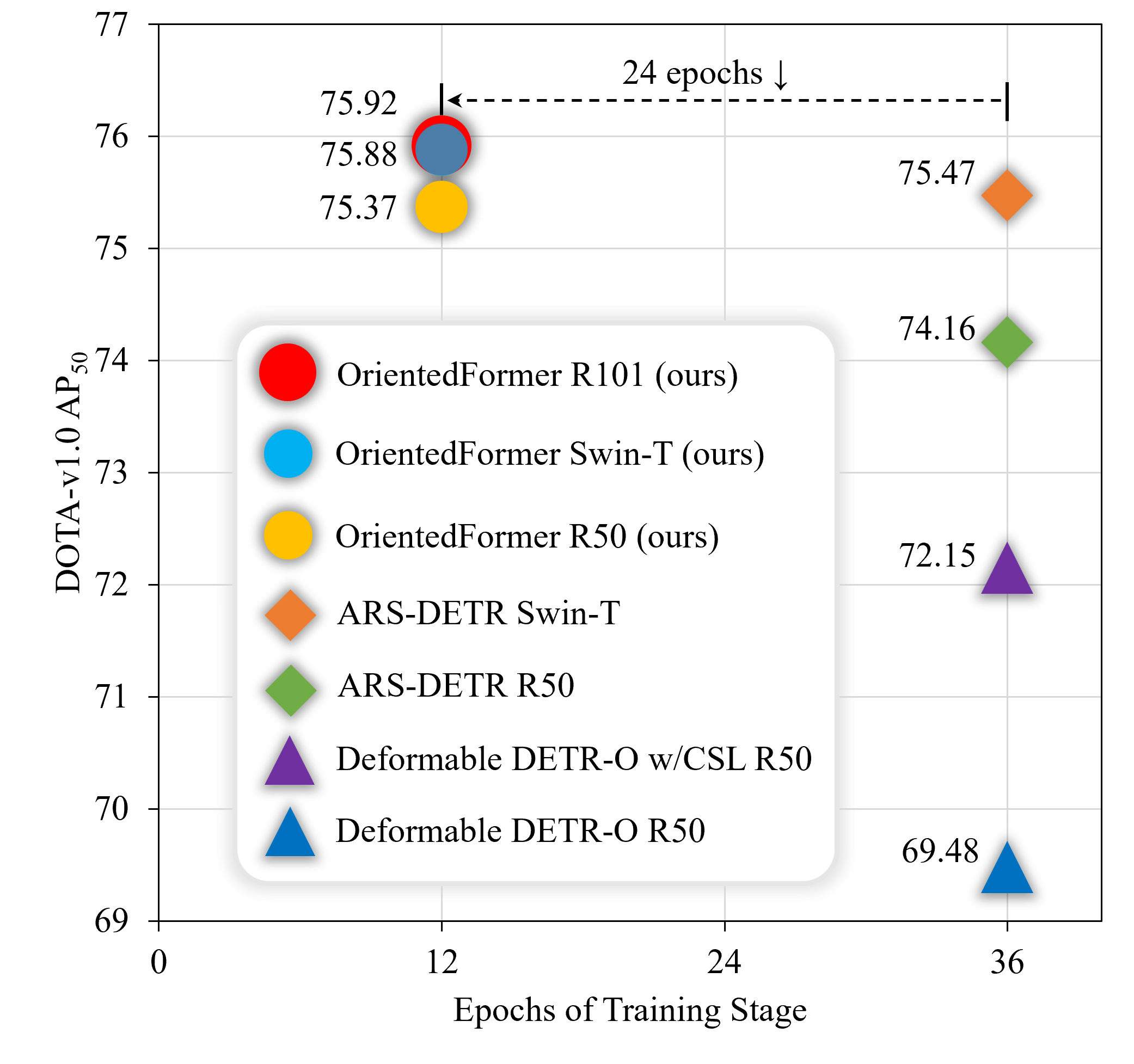}
       \caption{\textbf{Epochs} of training stage versus \textbf{accuracy} on the DOTA-v1.0 test set.}
       \label{epochs_vs_map}
    \end{figure}
}

\newcommand{\prcurveiouzeropointfivedior}{
    \begin{figure}[!t]
      \centering
       \includegraphics[width=0.98\linewidth]{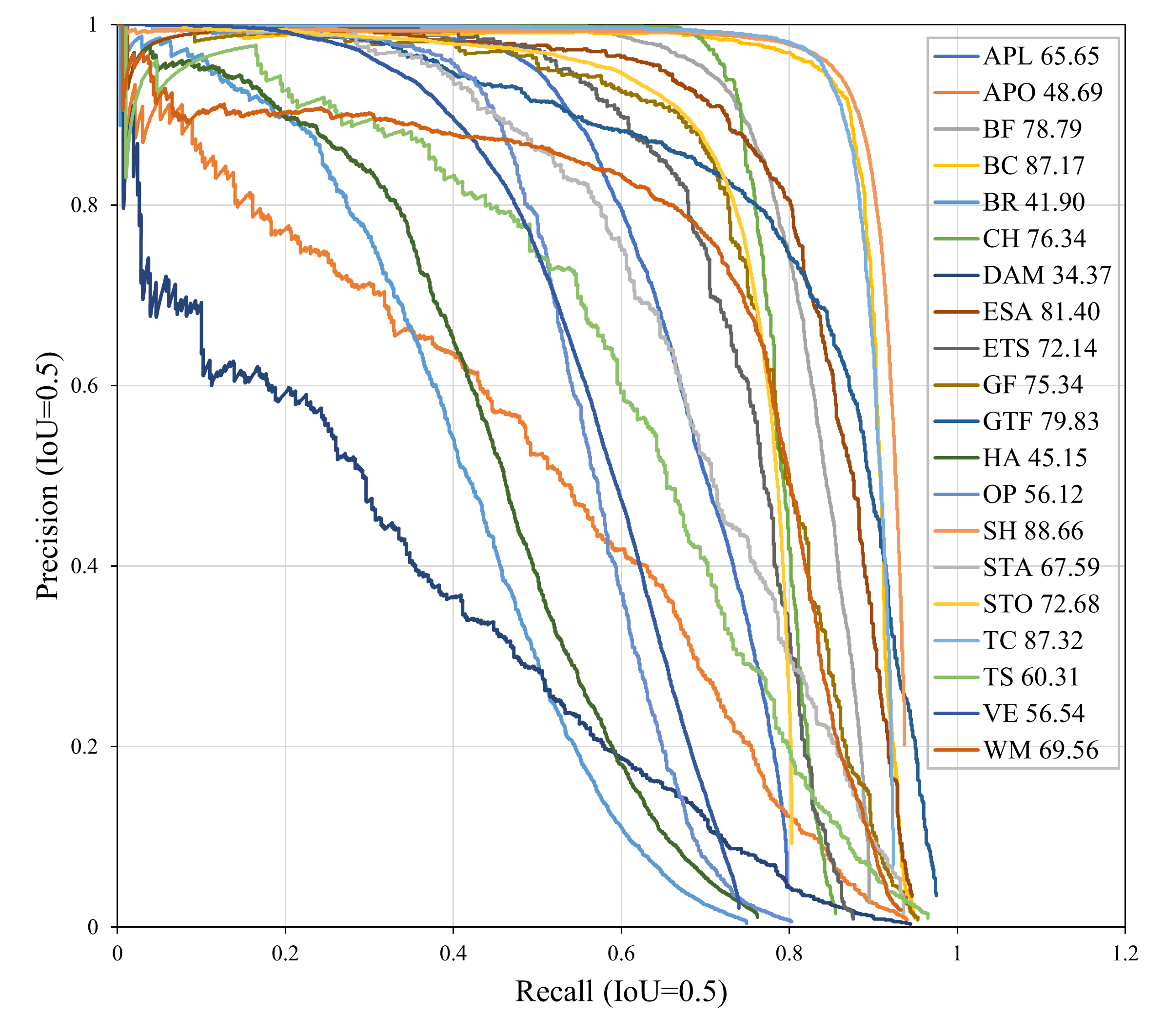}
       \caption{\textbf{P-R curve (IoU=0.5)} on the DIOR-R test set with ResNet50.}
       \label{pr_curve_iou0.5_dior}
    \end{figure}
}

\newcommand{\prcurveiouzeropointsevenfivedior}{
    \begin{figure}[!t]
      \centering
       \includegraphics[width=0.98\linewidth]{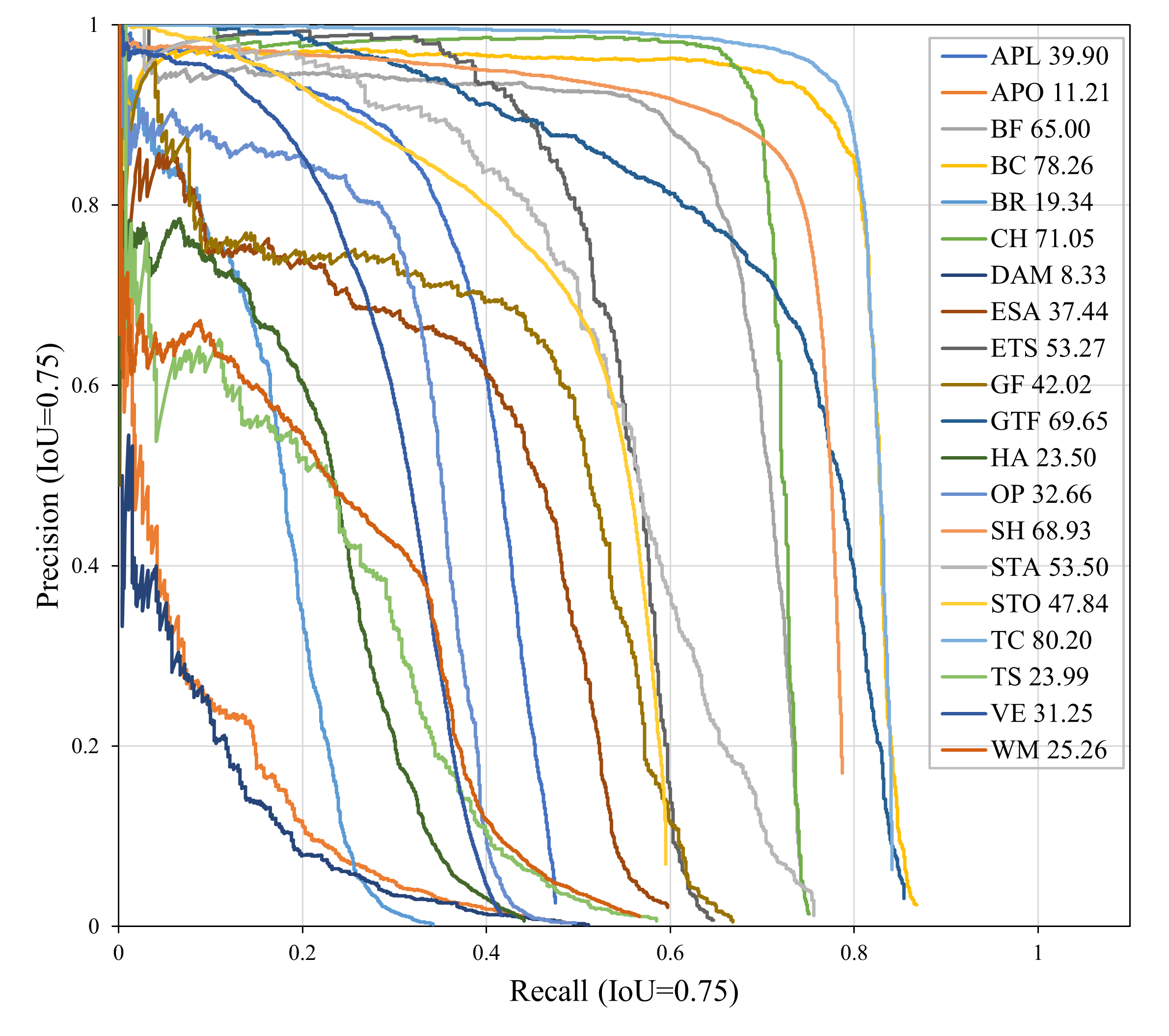}
       \caption{\textbf{P-R curve (IoU=0.75)} on the DIOR-R test set with ResNet50.}
       \label{pr_curve_iou0.75_dior}
    \end{figure}
}

\newcommand{\differentsamplingpoints}{
    \begin{figure}[!t]
      \centering
       \includegraphics[width=0.85\linewidth]{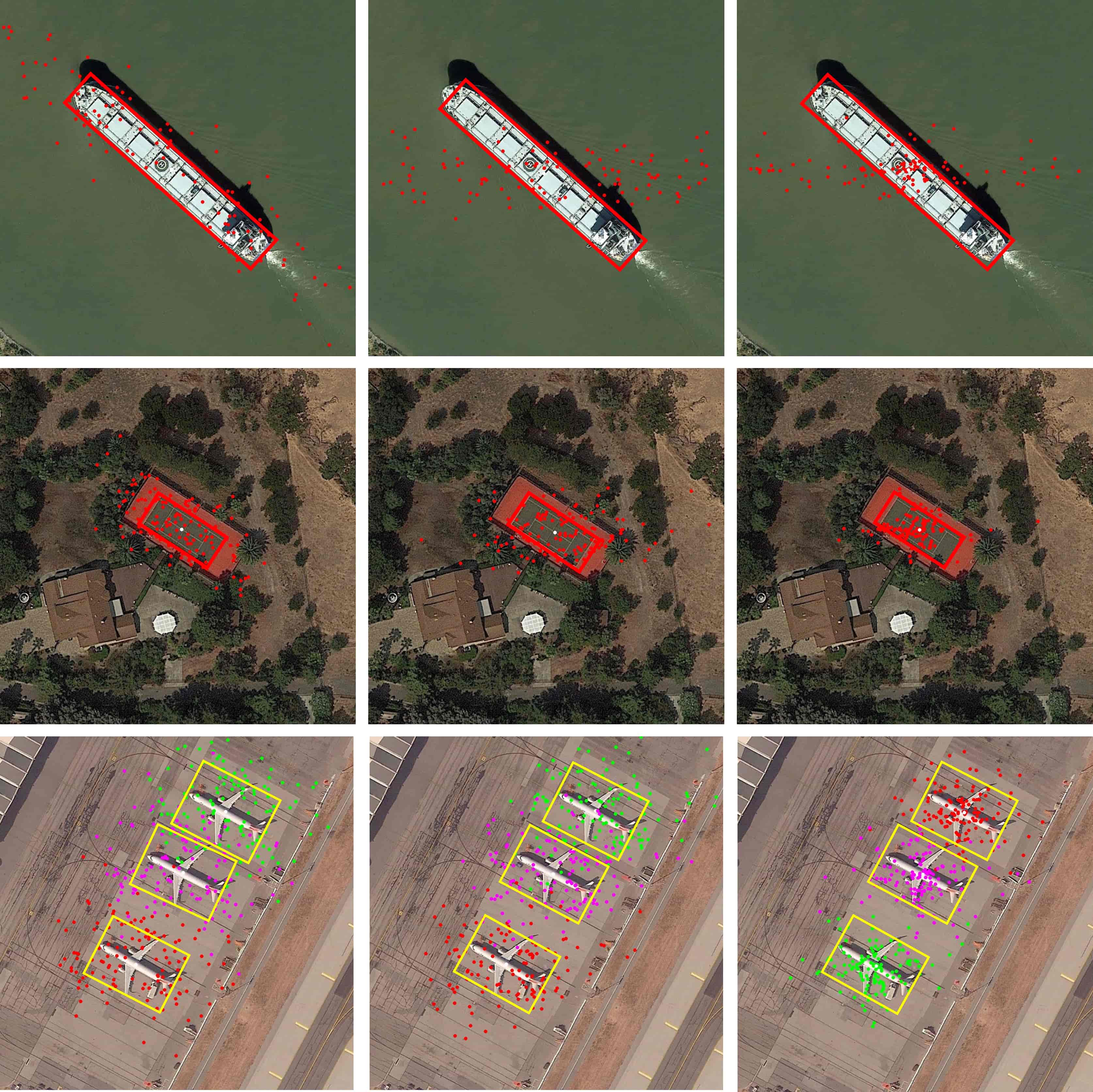}
       \caption{Sampling points in cross-attention. Oriented cross-attention (left column) rotates sampling points by angles for alignment. Deformable offsets method (middle column) does not rotate sampling points. Random offsets method (right column) employs random sampling points.}
       \label{differentsamplingpoints}
    \end{figure}
}

\newcommand{\suboptimalresults}{
    \begin{figure}[]
      \centering
       \includegraphics[width=0.96\linewidth]{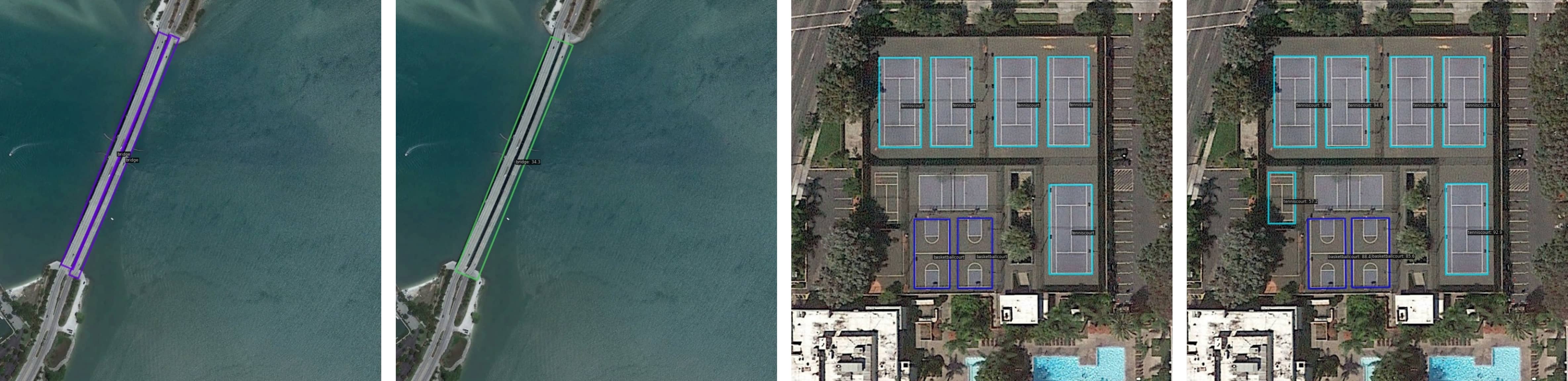}
       \caption{Suboptimal results.}
       \label{suboptimal results}
    \end{figure}
}

\newcommand{\querycenterpoints}{
    \begin{figure}[!t]
      \centering
       \includegraphics[width=0.85\linewidth]{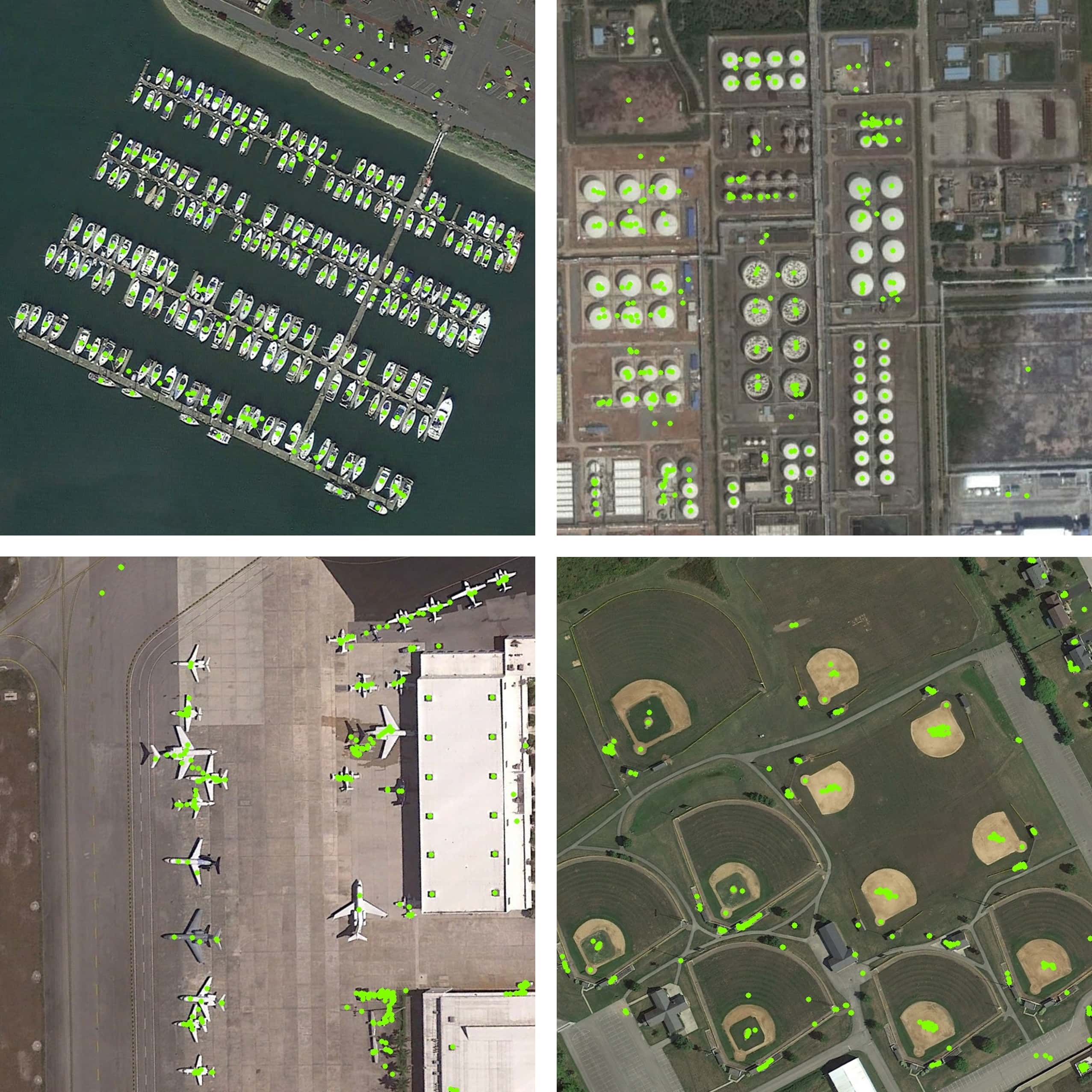}
       \caption{Center points of learned positonal queries.}
       \label{querycenterpoints}
    \end{figure}
}

\newcommand{\comparewithothermethods}{
    \begin{figure*}[!t]
    \centering
    \subfloat[OrientedFormer(ours)]
        {
            \includegraphics[width=0.175\linewidth]{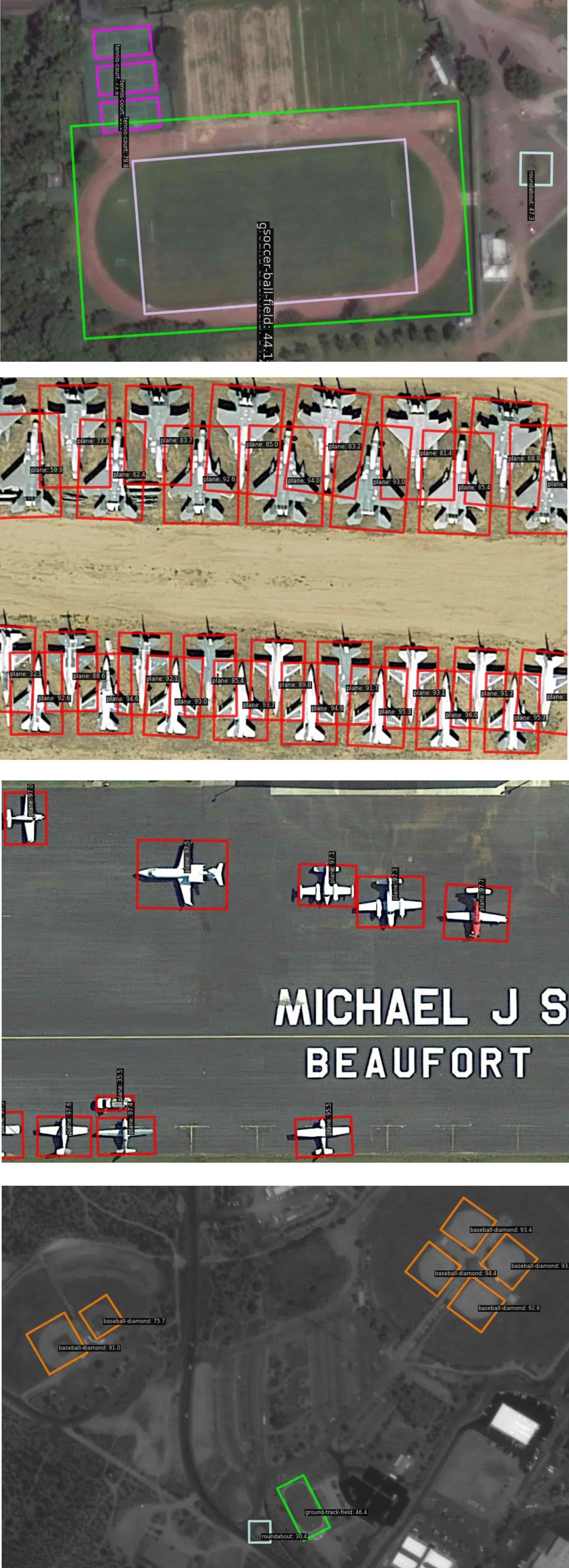}
            \label{compare_orientedformer}
        }
    \subfloat[Oriented R-CNN~\cite{orientedrcnn}]
        {
            \includegraphics[width=0.175\linewidth]{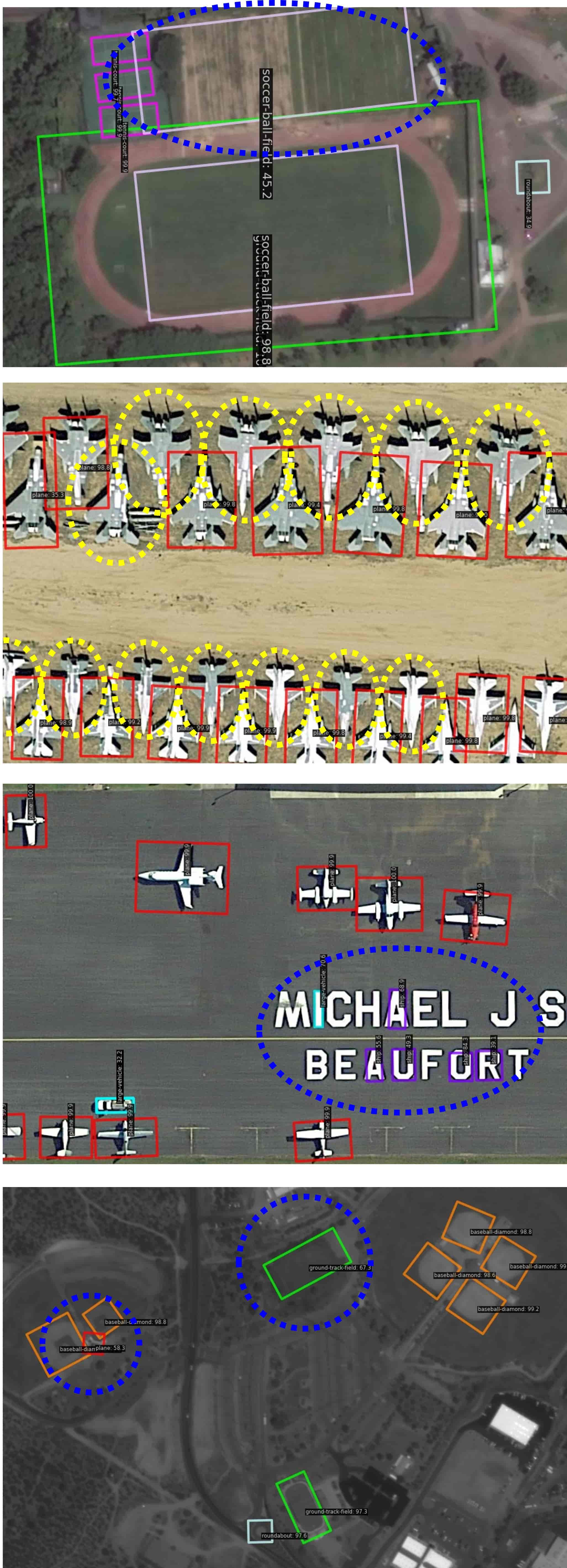}
            \label{compare_orientedrcnn}
        }
    \subfloat[SASM~\cite{sasm}]
        {
            \includegraphics[width=0.175\linewidth]{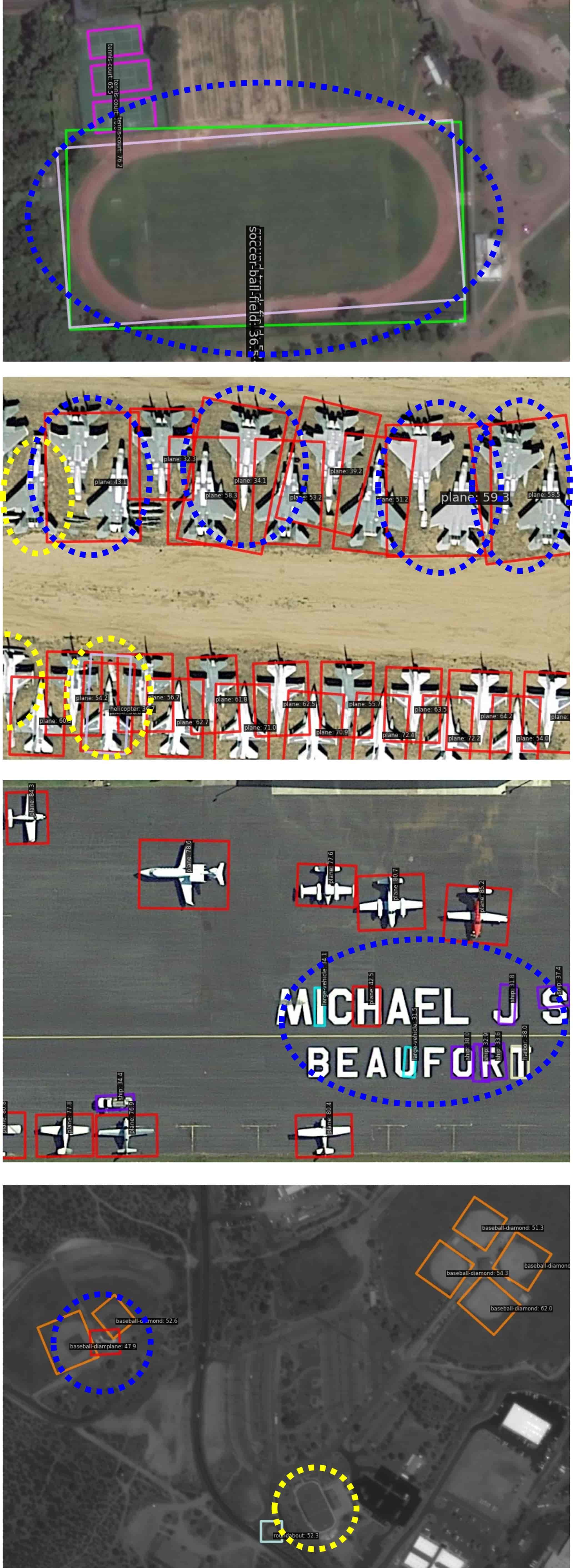}
            \label{compare_sasm}
        }
    \subfloat[ARS-DETR~\cite{arsdetr}]
        {
            \includegraphics[width=0.174\linewidth]{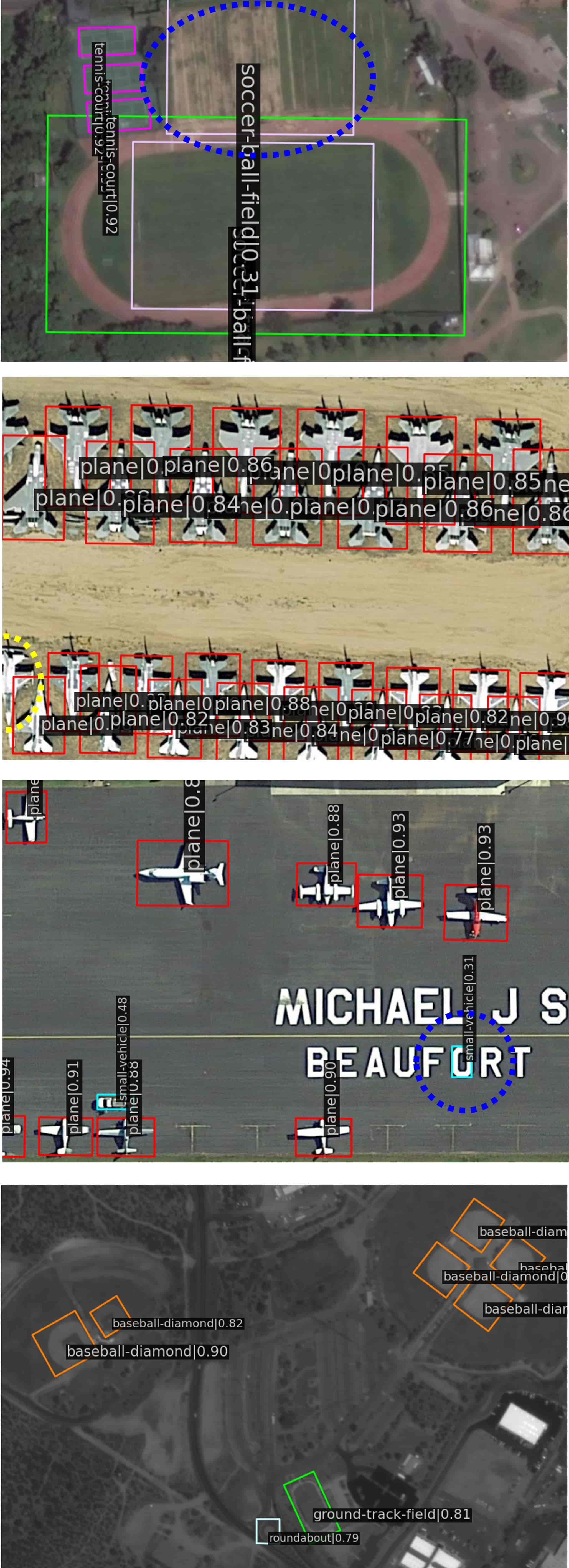}
            \label{compare_arsdetr}
        }
    \subfloat[Deformable DETR-O~\cite{deformabledetr}]
        {
            \includegraphics[width=0.175\linewidth]{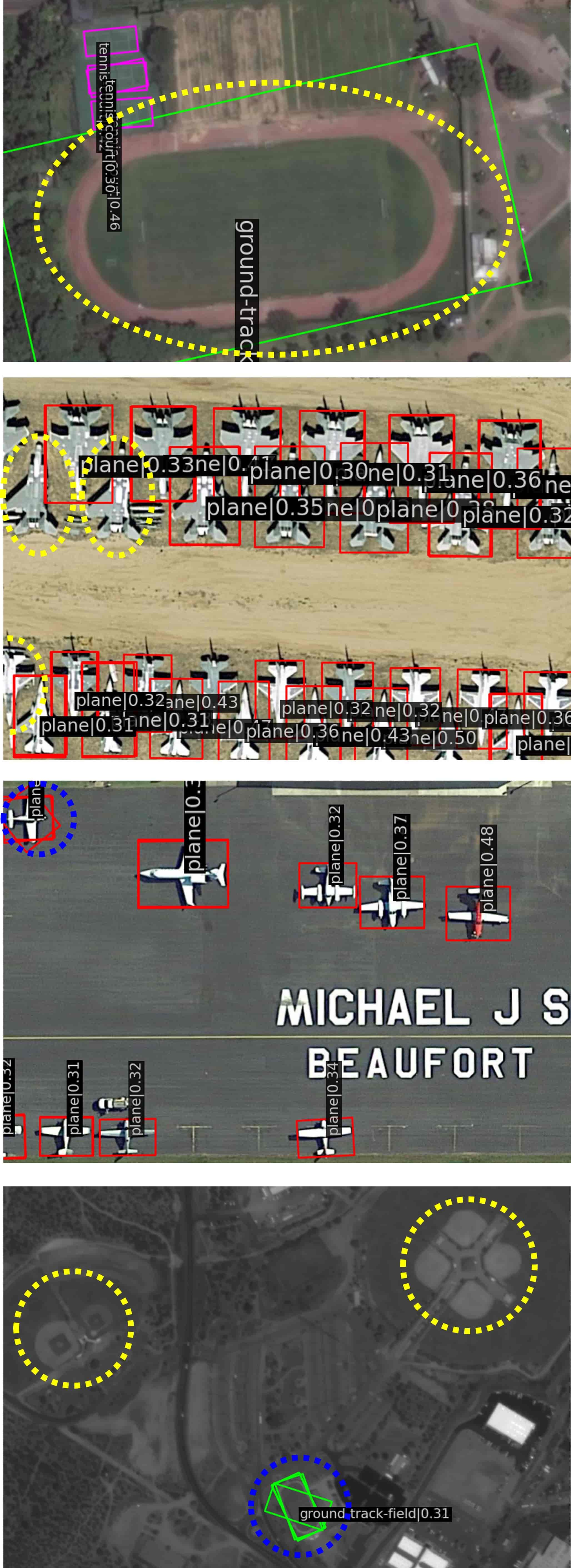}
            \label{compare_deformabledetr}
        }
    \caption{Comparison between our method and others on challenging samples. Confidence threshold is set to 0.3. Blue circles denote error results. Yellow circles denote missed results. The 1st line: Large-scale objects, e.g., soccer fields and ground track fields. The 2nd line: Densely packed objects, e.g., planes. The 3rd line: Images with complex backgrounds, e.g., the disturbance of letters. The 4th line: Images under poor environmental conditions.}
    \label{compare_with_other_method}
    \end{figure*}

}

\begin{document}

\title{OrientedFormer: An End-to-End Transformer-Based Oriented Object Detector in Remote Sensing Images}

\author{
Zeyu Ding\textsuperscript{~\orcidlink{0009-0005-9614-7131}}, 
Yong Zhou\textsuperscript{~\orcidlink{0000-0001-6207-0299}}, 
Jiaqi Zhao\textsuperscript{~\orcidlink{0000-0002-3564-5090}},~\IEEEmembership{Member,~IEEE}, 
Hancheng Zhu\textsuperscript{~\orcidlink{0000-0002-5418-9879}},
Wen-Liang Du\textsuperscript{~\orcidlink{0000-0002-9234-0912}}~\IEEEmembership{Member,~IEEE},\\ 
Rui Yao\textsuperscript{~\orcidlink{0000-0003-2734-915X}},~\IEEEmembership{Member,~IEEE}, 
and Abdulmotaleb El  Saddik\textsuperscript{~\orcidlink{0000-0002-7690-8547}},~\IEEEmembership{Fellow,~IEEE} \\
{\color{red}\faFire}\ ESI Highly Cited Paper

\thanks{Manuscript received 16 May 2024; revised 12 July 2024 and 12 August 2024; accepted 2 September 2024.}
\thanks{Zeyu Ding, Yong Zhou, Jiaqi Zhao, Hancheng Zhu, Wen-Liang Du, and Rui Yao are with the School of Computer Science and Technology/School of Artificial Intelligence, the Mine Digitization Engineering Research Center of the Ministry of Education, and Jiangsu Provincial Industrial Technology Engineering Center for Intelligent Sensing and Emergency IoT in Underground Space, China University of Mining and Technology, Xuzhou 221116, China (e-mails: dingzeyu@cumt.edu.cn, yzhou@cumt.edu.cn, jiaqizhao@cumt.edu.cn, zhuhancheng@cumt.edu.cn, wldu@cumt.edu.cn, ruiyao@cumt.edu.cn).}

\thanks{Abdulmotaleb El Saddik is with the School of Electrical Engineering and Computer Science, University of Ottawa, Ottawa, ON K1N 6N5, Canada (e-mail: elsaddik@uottawa.ca).}
\thanks{Digital Object Identifier 10.1109/TGRS.2024.3456240}
}

\markboth{IEEE TRANSACTIONS ON GEOSCIENCE AND REMOTE SENSING, VOL. 62, 2024}%
{Shell \MakeLowercase{\textit{et al.}}: A Sample Article Using IEEEtran.cls for IEEE Journals}


\maketitle

\begin{abstract}
Oriented object detection in remote sensing images is a challenging task due to objects being distributed in multi-orientation. Recently, end-to-end transformer-based methods have achieved success by eliminating the need for post-processing operators compared to traditional CNN-based methods. However, directly extending transformers to oriented object detection presents three main issues: 1) objects rotate arbitrarily, necessitating the encoding of angles along with position and size; 2) the geometric relations of oriented objects are lacking in self-attention, due to the absence of interaction between content and positional queries; and 3) oriented objects cause misalignment, mainly between values and positional queries in cross-attention, making accurate classification and localization difficult. In this paper, we propose an end-to-end transformer-based oriented object detector, consisting of three dedicated modules to address these issues. First, Gaussian positional encoding is proposed to encode the angle, position, and size of oriented boxes using Gaussian distributions. Second, Wasserstein self-attention is proposed to introduce geometric relations and facilitate interaction between content and positional queries by utilizing Gaussian Wasserstein distance scores. Third, oriented cross-attention is proposed to align values and positional queries by rotating sampling points around the positional query according to their angles. Experiments on six datasets DIOR-R, a series of DOTA, HRSC2016 and ICDAR2015 show the effectiveness of our approach. Compared with previous end-to-end detectors, the OrientedFormer gains 1.16 and 1.21 AP$_{50}$ on DIOR-R and DOTA-v1.0 respectively, while reducing training epochs from 3$\times$ to 1$\times$. Code is available at {\textcolor[HTML]{181717}{\raisebox{-0.25\height}{\simpleicon{github}}}} \url{https://github.com/wokaikaixinxin/ai4rs} and {\textcolor[HTML]{181717}{\raisebox{-0.25\height}{\simpleicon{github}}}} \url{https://github.com/wokaikaixinxin/OrientedFormer}.
\end{abstract}


\figureone

\section{Introduction}
\IEEEPARstart{O}{riented} object detection is a fundamental task in the intersection of computer vision and remote sensing, as it aims to locate objects by a set of oriented boxes and categorize them. Remote sensing images used in oriented object detection are photographs of target objects on the Earth's surface captured by satellites or other aerial platforms. Oriented object detection remains challenging due to objects being distributed with multiple orientations, dense arrangements, and varying scales, shown in Fig.~\ref{fig_first_case}. These characteristics of oriented objects make it difficult to localize and classify target objects accurately.

To detect objects accurately, oriented object detection methods~\cite{10008383, 9925528, 10380690} based on convolutional neural networks (CNNs) have made significant progress. Most of them are two-stage~\cite{orientedrcnn, redet, gliding_vertex} or one-stage~\cite{dcfl,sasm,oriented_reppoints} detectors. Two-stage methods select foreground proposal boxes using a region proposal algorithm in the first stage and refine these proposals in the second stage. For example, RoI Transformer learns to transform horizontal anchors into oriented anchors, but horizontal anchors often misalign with the instance features. To address this issue, RRPN generates abundant oriented proposals using a rotation region proposal network. Oriented R-CNN introduces midpoint offset representation in an oriented region proposal network. Meanwhile, one-stage detectors directly predict the location and category of anchor boxes to avoid complex proposal generation. For example, R3Det directly obtains oriented proposals and aligns features through refinement modules. However, the above CNN-based methods rely on a one-to-many label assignment strategy and require complex hand-designed post-processing operators, such as non-maximum suppression (NMS).

DETR~\cite{detr} first applies transformers~\cite{attention_is_all_you_need} to horizontal object detection, and a series of related works~\cite{deformabledetr,adamixer} have achieved promising performance. Inspired by them, some detectors adapt transformers for oriented object detection, typically following the encoder-decoder architecture. Compared to methods that adopt dense priors (e.g., boxes and points), transformer-based methods use a set of queries (e.g., content and positional queries) to represent object instances, which are usually updated layer by layer. The architecture of transformers mainly consists of three modules, positional encoding, self-attention, and cross-attention. Positional encoding is used to learn the sequence ordering of input tokens. The self-attention handles pairwise interactions between queries and removes duplicate predictions. Cross-attention facilitates interactions between values and queries, enabling models to focus on related regions. Furthermore, transformer-based detectors benefit from one-to-one label assignment and an end-to-end framework, which eliminates the need for complex hand-designed processes.

However, extending the transformer framework to oriented object detection presents three main issues that need to be overcome.
(1) \textit{\textbf{Objects rotate arbitrarily, requiring the encoding of angles}} in addition to position and size. Angles are used to characterize oriented objects, which distinguishes them from horizontal boxes. Angles, position, and size are all necessary to represent oriented objects. Current transformer-based methods~\cite{ao2detr} encode only position and size using vanilla positional encoding, but neglect angles. Additionally, we notice that the physical meanings and data ranges of angle $\theta$, position $(x,y)$, and size $(w, h)$ are different. Without normalization, the value range for coordinates and sizes is bounded by the size of images, whereas angles are measured in radians, ranging from $\left[-\pi/2, \pi/2\right]$ for the rotation of oriented boxes. Therefore, vanilla positional encoding is not suitable for oriented objects.
(2) \textit{\textbf{The geometric relations of oriented objects are lacking in self-attention.}} Content queries and positional queries represent semantic contextual features and spatial positions, respectively. Current transformer-based methods feed only content queries into self-attention, allowing them to interact with each other but failing to incorporate the geometric relation information provided by positional queries. Vanilla self-attention does not capture the geometric relationships between content queries.
(3) \textit{\textbf{Oriented objects cause misalignment.}} Objects rotate arbitrarily, while multi-scale image features have a pyramidal structure. This typically leads to misalignment between values and positional queries in cross-attention, where values are extracted from image features and positional queries represent boxes of oriented objects. This makes it difficult to accurately classify and localize target objects.

In this paper, we aim to alleviate the above issues for transformer-based oriented object detection. We propose an end-to-end transformer-based oriented object detection framework, called OrientedFormer. Our detector is equipped with three dedicated modules: Gaussian positional encoding (PE), Wasserstein self-attention, and oriented cross-attention. \textit{\textbf{First}}, for the issue of encoding of angles, the Gaussian PE is built on Gaussian distribution transformed from oriented boxes. It unifies angle, position, and size into the same metric and effectively encodes the angles of boxes. \textit{\textbf{Second}}, for the issue of lacking geometric relations, the Wasserstein self-attention enables content and positional queries to interact with each other. The geometric relation information is measured by Gaussian Wasserstein distance scores, and provided by all positional queries. \textit{\textbf{Third}}, for the issue of misalignment, the oriented cross-attention aligns values and positional queries. It rotates sparse sampling points around positional queries according to angles. The focused point regions distribute both inside and outside oriented boxes, which provides a wide range of contextual information, shown in Fig.~\ref{fig_second_case}. \textit{\textbf{Finally}}, we conduct extensive experiments on various oriented object detection datasets of remote sensing images. All experimental results consistently demonstrate the effectiveness of OrientedFormer in improving accuracy. Furthermore, we validated the generalization of our method by conducting studies on scene text detection.

In general, the contribution of our methods is summarized in four folds:
\begin{itemize}

\item[1)] The Gaussian positional encoding is proposed to encode the angle of oriented boxes in addition to position and size. It is constructed on Gaussian distributions, which unify angle, position, and size into the same metric.
\item[2)] The Wasserstein self-attention is proposed to introduce geometric relations into self-attention. This module utilizes Gaussian Wasserstein distance scores to measure the geometric relations between two different content queries.
\item[3)] To address the issue of misalignment, the oriented cross-attention is proposed to align values and positional queries by rotating a small set of sampling points around a positional query.
\item[4)] Extensive experiments on 6 datasets demonstrate the effectiveness of our approach. With the Resnet50 backbone, OrientedFormer achieves an AP$_{50}$ of 67.28\% on DIOR-R and 54.27\% on DOTA-v2.0 respectively, establishing new state-of-the-art benchmarks.
\end{itemize}

\section{Related Work}

\subsection{Oriented Object Detection in Remote Sensing}
\textbf{\textit{1) Convolution neural network (CNN) methods}} have achieved significant improvements in performance. Existing CNN-based oriented object detectors are mainly categorized into one-stage and two-stage methods. One-stage detectors predict the location and category of anchor boxes, which densely cover image feature maps, such as R3Det~\cite{r3det}, S$^{2}$-Net~\cite{s2anet}, and PSC~\cite{psc}. Anchor-free methods~\cite{cfa, oriented_reppoints} simplify the one-stage pipeline by replacing hand-crafted anchor boxes with prior points. One-stage methods rely on dense candidates, with each candidate supervised directly by classification and regression losses. During training, one-to-many assignment strategies based on pre-defined principles are used, such as whether the intersection-over-union (IoU) between candidates and ground truth boxes exceeds a threshold. Two-stage methods select foreground proposal boxes from dense region candidates in the first stage and localize and classify these proposal boxes in the second stage. The region proposal algorithm is used in the first stage to learn coarse proposal boxes, such as Region Proposal Networks (RPN) in Oriented R-CNN~\cite{orientedrcnn} and ReDet~\cite{redet}. Similar label assignment strategies are required in two-stage methods. Some post-processing operators, e.g., NMS are needed to remove redundant prediction results during inference time both in one-stage and two-stage methods.

\textbf{\textit{2) End-to-end Transformer-based methods}} are widely used in horizontal object detection~\cite{detr, deformabledetr}, which directly output the results without hand-crafted components. Some researchers~\cite{cledet} have extended them to oriented object detection. AO2-DETR~\cite{ao2detr} introduces an oriented boxes generation and refinement module for accurately oriented positional priors, building upon Deformable DETR~\cite{deformabledetr}. ARS-DETR~\cite{arsdetr} improves on previous work by proposing rotated deformable attention, wherein sampling points are rotated based on angles for feature alignment. In addition, certain methods focus on improving object queries. PSD-SQ~\cite{psd-sq} represents object queries as point sets instead of oriented boxes to enable accurate instance feature sampling. D$^2$Q-DETR~\cite{d2qdetr} designs dynamic queries that gradually reduce the number of object queries in the stacked decoder layers, aiming to better balance model precision and efficiency. Furthermore, several studies have concentrated on enhancing one-to-one label assignment. EMO2-DETR~\cite{emo2detr} observes and addresses the issue that one-to-one label assignment results in relative redundancy of object queries because objects are unevenly distributed in images. Different from existing approaches, to effectively encode oriented boxes, measure the geometric relations between content queries, and align values and positional queries, we propose the proposed Gaussian PE, Wasserstein self-attention, and oriented cross-attention.

\subsection{Attention in Transformer-based Object Detection}

\textbf{\textit{1) Self-attention:}} Object queries are fed into self-attention and interact with each other to remove duplicate predictions~\cite{detr}. Most transformer-based detectors adopt vanilla self-attention following DETR~\cite{detr}. In vanilla self-attention, only content queries are used, and the geometric relations provided by positional queries are lost. As a result, vanilla self-attention does not account for geometric relations between content queries. In contrast to this approach, we introduce Gaussian Wasserstein distance scores into self-attention to measure the geometric relations between different content queries.

\textbf{\textit{2) Cross-attention:}} In cross-attention, image features serve as values and interact with queries. The vanilla cross-attention in DETR~\cite{detr} only adopts a single feature map which is low efficiency. To accelerate the converging speed, deformable attention proposed in Deformable DETR~\cite{deformabledetr} attends to a small set of sampling points around a reference. Features corresponding to these points learn to classification and regression. But at the supervision of angles, sampling points learn to locate at special positions~\cite{ao2detr}, e.g. the catercorner and axis with boxes, which may be sub-optimal. Anchor DETR~\cite{anchor_detr} decouples attention into row and column attention and process them successively. The row and column sequences of image features lack orientation and spatial information. The SMCA~\cite{smca} proposes spatially modulated co-attention by constraining attention responses to be high near initially estimated bounding box locations. Dynamic DETR~\cite{dynamic_detr} designs RoI-based dynamic attention inspired by dynamic convolution~\cite{dynamic_conv} to assist the transformer in focusing on the region of interest. The above methods will cause misalignment when facing oriented boxes. Different from them, our oriented cross-attention aligns values and positional queries by rotating a small set of sampling points around a positional query.

\subsection{Positional Encoding}
Positional encoding is important for transformers to capture the sequence ordering of input tokens. It is first applied in the transformer~\cite{attention_is_all_you_need} to inject information about the relative or absolute~\cite{attention_is_all_you_need} positional of tokens in a sequence for natural language processing. Since the transformer does not have convolution, it needs positional encoding to learn the sequence ordering of tokens. The above methods are designed for 1D word in sequence language models, and beyond that, positional encoding is widely employed in computer vision. In object detection, DETR~\cite{detr} employs learnable positional encoding. The positional encoding in DAB DETR~\cite{dab_detr} maps center coordinates $(x,y)$ and size $(w, h)$ of boxes to four vectors respectively, and contacts them as final embeddings. The variant two-stage Deformable DETR~\cite{deformabledetr} generates region proposals and then encodes them by sinusoidal absolute positional encoding. The above methods only encode horizontal boxes. Different from them, our proposed Gaussian positional encoding is constructed on Gaussian distribution, which is transformed from oriented boxes. It can encode the angles, position, and size of oriented boxes.

\subsection{Nomenclature}
To facilitate clarity in the subsequent discussion, we present a summary of symbols along with their corresponding descriptions as utilized in this study, encapsulated in Table~\ref{notations}.
%We present a summary of symbols along with descriptions as utilized in this study, encapsulated in Table~\ref{notations}.

\notations
\overallarchitecture

\section{Method}
In this paper, we propose OrientedFormer, an end-to-end transformer-based oriented object detector in remote sensing images. In this part, we first introduce the overall architecture in Sec.~\ref{section_overall_architecture}, and then illustrate object queries in Sec.~\ref{section_object_queries}, Gaussian positional encoding in Sec.~\ref{section_gaussian-like_pe}, Wasserstein self-attention in Sec.~\ref{section_gaussian-like_self_attn} and oriented cross-attention in Sec.~\ref{section_oriented_3d_attn}, respectively. Lastly, label assignments and losses are introduced in Sec.~\ref{section_update_label_assignment_and_loss}.

\subsection{Overall Architecture}
\label{section_overall_architecture}
In general, the architecture of our OrientedFormer is composed of a backbone and a decoder, which follows the paradigm of end-to-end transformers, shown in Fig.~\ref{fig_overall_framework}. The encoder is not used and queries are initialized by enhancement method~\cite{ddq}. Multi-scale image features $\left\{f^l\right\}^L_{l=1}$ are extracted by the backbone~\cite{resnet} and transformed in the same channel via the channel mapper~\cite{deformabledetr}, where $f^l\in\mathbb{R}^{D\times H_l\times W_l}$ and $l\in\left\{ 1,2,...,L\right\}$ denote a single level feature and different scales respectively. A single-level feature $f^l$ is with channel $D$ (256 by default), height $H_l$, and width $W_l$. The downsampling stride between two adjacent features is usually 2. Multi-scale features and object queries are the inputs of the decoder. Following~\cite{detr, deformabledetr}, we sequentially use our proposed self-attention, cross-attention, and feedforward-feed network (FFN) in the decoder. In self-attention, object queries interact with each other, while in cross-attention, sampled features as values further interact with queries. Through FFN, updated queries and detection results are produced. During training times, predictions are supervised by classification and regression losses.

\subsection{Object Queries}
\label{section_object_queries}
Object queries are one of the inputs of the decoder for representations of object instances. Content queries $Q_c \in \mathbb{R}^{N\times D}$ and positional queries $Q_p \in \mathbb{R}^{N\times 5}$ are used to learn contextual image information and positions of objects respectively, where $N$ denotes the number of object queries and $D$ (256 by default) denotes channel dimension. These two kinds of queries disentangle the classification and localization of objects.

\subsection{Gaussian positional encoding}
\label{section_gaussian-like_pe}
The modern PE is exclusively employed for horizontal boxes and inaccurately encodes angles of oriented objects. To address the limitations, we propose Gaussian PE, which can encode angle, position, and size uniformly.

\textbf{\textit{1) Preliminaries of PE:}} In the decoder, positional encoding transforms positional queries into sinusoidal embeddings, and then content queries are trained jointly with positional embeddings. We review modern positional encoding in object detection first. In many common horizontal object detectors~\cite{deformabledetr, adamixer}, positional encoding applied to queries can be formulated as:
\begin{equation}
\varphi(\mathbf{x})=[\sin(\mathbf{x}),\cos(\mathbf{x}),...,\sin(\frac{\mathbf{x}}{T^{\frac{2K}{D'}}}),\cos(\frac{\mathbf{x}}{T^{\frac{2K}{D'}}})]^{\top}
    \label{original_pe}
\end{equation}
This is the concatenation of sine and cosine values of each dimension of the horizontal boxes $\mathbf{x}$ $(x, y, w, h)$ scaled by $1/{T^{\frac{2k}{D'}}}$, where $T$, $k\in \{0,1,...,K\}$, $D'$ are temperature, hyperparameter, and dimension respectively.

\textbf{\textit{2) Gaussian Distribution of Oriented Boxes}}: To unify angle, position, and size into the same metric, we convert an oriented box into a Gaussian distribution $\mathcal{N}(\boldsymbol{\mu},\space \boldsymbol{\Sigma})$:
\begin{equation}
    \begin{split}
\boldsymbol{\mu}&=(x,y),\\
\mathbf{\Sigma}&=\mathbf{R}\mathbf{\Lambda}\mathbf{R}^{\top}\\
&=\begin{pmatrix}
  \cos\theta &-\sin\theta  \\
  \sin\theta &\cos \theta
\end{pmatrix}
\begin{pmatrix}
  \frac{w^2}{4} &0 \\
  0&\frac{h^2}{4}
\end{pmatrix}
\begin{pmatrix}
  \cos \theta &\sin\theta  \\
  -\sin\theta &\cos \theta
\end{pmatrix}
\end{split}
    \label{gau_rbox}
\end{equation}
where $\mathbf{R}$ is the 2D rotation matrix, and $\mathbf{\Lambda}$ is the diagonal matrix of eigenvalues.

\gaussianlikepe

\textbf{\textit{3) Gaussian PE:}} The proposed Gaussian PE is the expectation of oriented boxes distributed according to the aforementioned Gaussian. The original positional encoding in Eq. (\ref{original_pe}) can be rewrite as:
\begin{equation}
\varphi(\mathbf{x})_{2i}=\sin (T^{-\frac{2i}{D'}}\cdot\mathbf{x}),\quad\varphi(\mathbf{x})_{2i+1}=\cos (T^{-\frac{2i}{D'}}\cdot\mathbf{x})
\end{equation}
where the superscript $2i$ and $2i+1$ denote the indices in the encoded vectors. This reparameterization makes it possible to derive a closed form for Gaussian PE. There are two mathematical facts the expectation of a linear transformation of random variables is a linear transformation of the random variables' expectation, and the variance of a linear transformation of random variables is the product of variance and the square of coefficient. According to these properties, we can calculate the mean and covariance of Gaussian distributions of oriented boxes after they are lifted for positional encoding:
\begin{equation}
\boldsymbol{\mu}_{\varphi}=T^{-\frac{2i}{D'}}\cdot\boldsymbol{\mu},\enspace \mathbf{\Sigma}_{\varphi}=(T^{-\frac{2i}{D'}})^2\cdot\mathbf{\Sigma}
    \label{lifted_gau}
\end{equation}

The last step of our Gaussian PE is calculating expectations over the lifted multivariate Gaussian in Eq. (\ref{lifted_gau}), which is modulated by sinusoidal and cosine functions. There is another mathematic fact that if $x$ distributes in Gaussian with mean $\mu$ and variance $\sigma$, the expectation value $\mathrm{E}[\sin(x)]$ and $\mathrm{E}[\cos(x)]$ are~\cite{mip_nerf}:
\begin{equation}
\begin{split}
    \mathrm{E}_{x\sim\mathcal{N}(\mu,\sigma^{2})}[\sin(x)]&=\sin(\mu)\exp\bigl(-(\nicefrac{1}{2})\sigma^{2}\bigr), \\
    \mathrm{E}_{x\sim\mathcal{N}(\mu,\sigma^{2})}[\cos(x)]&=\cos(\mu)\exp\bigl(-(\nicefrac{1}{2})\sigma^{2}\bigr)
\end{split}
    \label{expectation_of_sin}
\end{equation}
as we can see, the mathematic expectation $\mathrm{E}[\sin(x)]$ and $\mathrm{E}[\cos(x)]$ are the $\sin(\cdot)$ and $\cos(\cdot)$ of mean $\mu$ attenuated by $\exp(\cdot)$ of the variance $\sigma$. With the property, the proposed Gaussian PE is calculated as:
\begin{equation}
\begin{aligned}
\varphi (\boldsymbol{\mu},\boldsymbol{\Sigma})& =\mathrm{E}_{\mathbf{x}\sim\mathcal{N}(\boldsymbol{\mu}_{\varphi},\boldsymbol{\Sigma}_{\varphi })}\left[\varphi(\mathbf{x})\right]  \\
&=\begin{bmatrix}\sin(\boldsymbol{\mu}_\varphi )\circ\exp(-(\nicefrac{1}{2})\mathrm{diag}(\boldsymbol{\Sigma}_\varphi))\\\cos(\boldsymbol{\mu}_\varphi)\circ\exp(-(\nicefrac{1}{2})\mathrm{diag}(\boldsymbol{\Sigma}_\varphi))\end{bmatrix}
\end{aligned}
    \label{final_gau_pe}
\end{equation}
where $\circ$ denotes the hadmard product. Since positional encoding encodes each dimension of boxes $\mathbf{x}$ independently, the Gaussian positional encoding depends on only the marginal distribution of $\varphi(\mathbf{x})$. Thus, only the diagonal of the covariance matrix $\boldsymbol{\Sigma}_{\varphi}$ is required:
\begin{equation}
    \mathrm{diag}(\boldsymbol{\Sigma}_\varphi)=[\mathrm{diag}(\boldsymbol{\Sigma}),...,(\frac{1}{T^{2K/D'}})^2\mathrm{diag}(\boldsymbol{\Sigma})]^\top
    \label{diag_of_var}
\end{equation}
These diagonals can be easily obtained from the variance $\boldsymbol{\Sigma}$ of Gaussian distribution of oriented boxes in Eq. (\ref{gau_rbox}).

%The positional embeddings of our proposed Gaussian PE are added to content queries and then they are fed into self-attention:
%\begin{equation}
%    \begin{split}
%&\mathrm{Attn}(Q_{c},\varphi)= \\
%&\mathrm{Softmax}((Q_{c}+\varphi(\boldsymbol{\mu},\boldsymbol{\Sigma}))(Q_{c}+\varphi(\boldsymbol{\mu},\boldsymbol{\Sigma}))^\top/\sqrt{d_{q}})Q_{c}
%\end{split}
%\end{equation}
%where $d_q$ denotes dimensions.

\gaussianlikeselfattn
\orientedthreedattn

\subsection{Multi-head Wasserstein Self-attention in Decoder}

\label{section_gaussian-like_self_attn}
The vanilla multi-head self-attention~\cite{detr} used between content queries does not account for geometric relation information. To address this, we propose Wasserstein self-attention, which introduces geometric relations into self-attention mechanisms and can effectively suppress redundant detections~\cite{detr}.

\textbf{\textit{1) Wasserstein Self-attention:}} We introduce Gaussian Wasserstein distance scores into the self-attention. It can measure the geometric relations between two different queries and assist self-attention focus on important areas, which is the main difference between Wasserstein self-attention and other vanilla self-attention. Given any two positional queries $\mathbf{x}_1$, $\mathbf{x}_2$ with Gaussian distributions $\mathbf{x}_1\sim\mathcal{N}(\boldsymbol{\mu}_1,\boldsymbol{\Sigma}_1)$ and $\mathbf{x}_2\sim\mathcal{N}(\boldsymbol{\mu}_2,\boldsymbol{\Sigma}_2)$ in Eq. (\ref{gau_rbox}), the Wasserstein distance is calculated as:
\begin{equation}
    d^2_{ij}=\left \|\boldsymbol{\mu}_1-\boldsymbol{\mu}_2\right \| ^2_2+\mathrm{Tr}(\boldsymbol{\Sigma}_1+\boldsymbol{\Sigma}_2-2(\boldsymbol{\Sigma}_1^{\nicefrac{1}{2}}\boldsymbol{\Sigma}_2\boldsymbol{\Sigma}_1^{\nicefrac{1}{2}})^{\nicefrac{1}{2}})
    \label{wasserstein_distance}
\end{equation}
where $i, j\in\left\{1,2,...,N\right\}$ and $i,\enspace j$ denote two Gaussian distributions of any two positional queries. The distance only satisfies $\ge0$. Thus we further rescale it and finally get the Gaussian Wasserstein distance scores:
\begin{equation}
    G_{ij}=\log(\frac{1}{\tau+d_{ij}}+\epsilon)
    \label{gwd_scores}
\end{equation}
where $\tau=1$, $\epsilon=10^{-7}$ and $i,j\in\left\{1,2,...,N\right\}$. The $G_{ij}=0$ stands for the positional queries $i$ and $j$ coinciding and $G_{ij}=\log\epsilon\ll0$ indicates that the two positional queries $i$ and $j$ are far away from each other. The proposed multi-head Wasserstein self-attention combines Gaussian PE and Gaussian Wasserstein distance scores:
\begin{equation}
    \begin{split}
&\mathrm{WAttn}(Q_{c},\varphi,G)= \\
&\mathrm{Softmax}((Q_{c}+\varphi(\boldsymbol{\mu},\boldsymbol{\Sigma}))(Q_{c}+\varphi(\boldsymbol{\mu},\boldsymbol{\Sigma}))^\top/\sqrt{d_{q}}+ G)Q_{c}
\end{split}
\end{equation}

\textbf{\textit{2) Complexity of Wasserstein Self-attention:}} The complexity of Wasserstein self-attention is $\mathcal{O}(D^2N+DN^2)$, on the same order of magnitude as other methods~\cite{detr,arsdetr}. The Gaussian Wasserstein distance scores and Gaussian PE do not impose additional computational burden.

\subsection{Multi-head oriented cross-attention in Decoder}
\label{section_oriented_3d_attn}
We propose oriented cross-attention for the issue of misalignment, shown in Fig.~\ref{fig_oriented_3d_attn}. The inputs of oriented cross-attention contain multi-scale image features $\left\{f^l\right\}^L_{l=1}$, content queries $Q_c\in\mathbb{R}^{N\times D}$ and positional queries $Q_p\in \mathbb{R}^{N\times 5}$, where $f^l$, $l$ denote a single level feature and different scales respectively. Oriented cross-attention can be analyzed from three perspectives: (1) Positional queries are transformed into another type $(x, y, z, r, \theta)$, which provide virtual 3D coordinates; (2) Values are sampled from image features, and sampling points are rotated according to angles for alignment; (3) The proposed cross-attention can be decoupled into three different attention mechanisms, each focusing on a different perspective: scale-aware attention, spatial-aware attention, and channel-aware attention.

The differences between oriented cross-attention and deformable attention are as follows: (1) our attention rotates sampling points according to angles for alignment, while deformable attention does not; (2) our attention focuses on three perspectives: scale-aware, spatial-aware and channel-aware attention, while deformable attention only emphasizes channel-awareness; (3) our attention utilizes learnable positional queries, whereas deformable attention employs a set of fixed mesh grids as reference points; (4) sampling points in our attention are distributed in a virtual 3D feature space, while in deformable attention, they are confined to a 2D plane.

\textbf{\textit{1) Coordinates of positional queries:}} We transform positional queries into another type $(x, y, z, r, \theta)\in \mathbb{R}^5$, where $(x, y)$ indicates coordinates of the center point, $(z, r)$ denotes the logarithm of scale and aspect ratio and $\theta$ represents the angle. The only difference between $(x, y, z, r, \theta)$ and the common 5-parameter representation $(x, y, w, h, \theta)$ is $(w, h)$ and the conversion is as follows:
\begin{equation}
    z=\log_{2}{\sqrt{wh}},\enspace r=\log_{2}{\frac{h}{w}}
    \label{conversion}
\end{equation}
where $(w, h)$ denotes the width and height respectively of oriented boxes. The purpose is that $(x, y, z)$ represents $3D$ coordinates in the virtual $3D$ feature space. Benefiting from this, it is easy to achieve scale, spatial, and channel attention.

\textbf{\textit{2) Calculation of values and feature alignment:}}
We sample features as values. Sampling points can be obtained from offsets of the center $(x, y, z)$ of a positional query $Q_{p_i} (x, y, z, r, \theta)$. The offsets $(\Delta x, \Delta y, \Delta z)$ is calculated as:
\begin{equation}
    (\Delta x, \Delta y, \Delta z) =\mathrm{Linear}(Q_c)
    \label{offsets}
\end{equation}
These offsets are transformed into sampling points:
\begin{equation}
    \left\{
\begin{aligned}
\tilde{x}&=x+\Delta x\cdot2^{z-r}\\
\tilde{y}&=y+\Delta y\cdot2^{z+r}\\
\tilde{z}&=z+\Delta z
\end{aligned}
\right.
    \label{sampling_points}
\end{equation}
Like other popular cross-attention, we also introduce multiple heads in oriented cross-attention. Thus, the number of sampling points around a position query is $g\cdot O$, where $g$ and $O$ denote the number of heads and sampling points, respectively.

Because objects in remote sensing images are oriented, we need to align the sampling points according to angles $\theta$:
\begin{equation}
    P=\begin{pmatrix}
 \cos \theta &-\sin\theta  \\
 \sin\theta &\cos \theta
\end{pmatrix}
\begin{pmatrix}
 \tilde {x}\\
\tilde {y}
\end{pmatrix}
    \label{aglin_sampling_points}
\end{equation}
The $\tilde{z}$ does not participate in rotating, because oriented boxes only rotate in 2D planes. Our task is 2D oriented object detection, which can not obtain real depth information like other 3D tasks. Thus, we need to rescale sampling points to adapt to different levels of features. The parameter $\tilde{z}$ is not directly involved in feature sampling but will be transformed into attention weights in scale-aware attention, which will be elaborated below. Given the aligned sampling points $P$, we rescale them first, and then sample values $V$ by bilinear interpolation in every level of features $\left\{f^l\right\}^L_{l=1}$:
\begin{equation}
    V^l=\mathrm{interpolation}(f^l,P/s^l)
    \label{get_the_values}
\end{equation}
 where $s^l$ denotes the downsampling stride of each level of features, and $l\in\left\{1,2,...,L\right\}$. The aligned sampling points are not strictly restricted to boxes. The values $V^l$ now are of the shape $\mathbb{R}^{N\times g\times O\times(D/g)}$, because the multi-head mechanism is used and the number of sampling points around a position query is $g\cdot O$.

%------------scale aware attn-----------------
\textbf{\textit{3) Scale-aware attention:}} We introduce scale-aware attention which dynamically fuses features of different scales:
\begin{equation}
    \pi_L(Q_c,V)=\sum_{l=1}^{L} \mathrm{Sigmoid}(-(\tilde{z}-\log_2(s^l))^2/\eta)\cdot V^l
    \label{scale-aware_attn}
\end{equation}
where $s^l$ denotes the downsampling stride of each level of features, and $l\in\left\{1,2,...,L\right\}$. The $\eta$ is the softing coefficient for the weight over different scales and we keep $\eta=2$ in the work. We denote the output of scale-aware attention as $V_{\pi_L}$. The $V_{\pi_L}$ is of the shape $\mathbb{R}^{N\times g\times O\times(D/g)}$ and continuously used in the channel-aware attention.

%------------channel aware attn-----------------
\textbf{\textit{4) Channel-aware attention:}}
To pay attention to channel dimensions, we introduce channel-aware attention. Given the output of scale-aware attention $V_{\pi_L}$ and content queries $Q_c$, the channel-aware attention is calculated as:
\begin{equation}
    \pi_C(Q_c,V_{\pi_L})=\mathrm{Sigmoid}(\gamma_1(\rho(\gamma_1(Q_c)))\cdot V_{\pi_L}
    \label{channel-aware_attn}
\end{equation}
where $\gamma_1$, $\rho$ are Linear and ReLU operator respectively. The weights for the channel are transformed from content queries by Linear and ReLU operators, and have the shape of $\mathbb{R}^{N\times g\times 1\times(D/g)}$. We denote the output of channel-aware attention as $V_{\pi_C}$. The $V_{\pi_C}$ is of the shape $\mathbb{R}^{N\times g\times O\times(D/g)}$ and continuously used in the spatial-aware attention.

%------------spatial aware attn-----------------
\textbf{\textit{5) Spatial-aware attention:}}
To focus on spatial dimensions, we introduce spatial-aware attention. Given the output of channel-aware attention $V_{\pi_C}$ and content queries $Q_c$, the spatial-aware attention is calculated as:
\begin{equation}
    \pi_S(Q_c,V_{\pi_C})=\mathrm{Sigmoid}(\gamma_2(\rho(\gamma_2(Q_c)))\cdot V_{\pi_C}
    \label{spatial-aware_attn}
\end{equation}
where $\gamma_2$ is the Linear operator. The weights for spatial are also transformed from content queries by Linear and ReLU operators and have the shape of $\mathbb{R}^{N\times g\times O\times 1}$. The output of spatial-aware attention is of the shape $\mathbb{R}^{N\times g\times O\times(D/g)}$. The output is further flattened and transformed to $\mathbb{R}^{N\times D}$ by a linear layer as the final output of oriented cross-attention to add back to the content queries.

%\textbf{\textit{6) Complexity of oriented cross-attention:}} The complexity for computing sampling points in Eq. (\ref{offsets}) and Eq. (\ref{sampling_points}) is of $\mathcal{O}(gOND)$. Given sampling points, the complexity of computing values in Eq. (\ref{get_the_values}) is $\mathcal{O}(LOND)$. The complexity of scale-aware, channel-aware, and spatial-aware attention is $\mathcal{O}(LOND)$, $\mathcal{O}(OND+ND^2)$, and $\mathcal{O}(ND^2+gOND+OND)$, respectively. The complexity for projection of final output is $\mathcal{O}(OND^2)$. So the overall complexity of oriented cross-attention is $\mathcal{O}((G+L+D)OND+ND^2)$.

\Algorithmtrain

\subsection{Label Assignment and Loss}
\label{section_update_label_assignment_and_loss}
In oriented object detection tasks, there are two subtasks, classification for categories and regression for positions of objects. At the stage of label assignment, one-to-one Hungarian matching~\cite{hungarian_matching} is used. The losses consists of the Focal loss~\cite{focal_loss} for classification, $L_{1}$ loss and Rotate IoU loss~\cite{rotated_iou_loss} for regression:
\begin{equation}
    \mathcal{L}=\lambda_{cls}\mathcal{L}_{cls}+\lambda_{L_1}\mathcal{L}_{1}+\lambda_{iou}\mathcal{L}_{iou}
    \label{losses}
\end{equation}
where $\lambda_{cls}$, $\lambda_{L_{1}}$ and $\lambda_{iou}$ are coefficient of corresponding losses. The losses applied at the detection results of all the decoder layers for training.

\section{Experiment}

\subsection{Datasets}

We conduct our experiments on 6 common datasets.
% towards oriented object detection, including DIOR-R~\cite{dior}, DOTA-v1.0~\cite{dotav1.0}/v1.5/v2.0~\cite{dota-v2.0}, HRSC2016~\cite{hrsc2016} and ICDAR2015~\cite{icdar2015}.
\textbf{\textit{DIOR-R}}\cite{dior} is a large-scale oriented object detection dataset for remote sensing images. It consists of 23,463 images and 192,512 instances that belong to 20 common classes. We train our model on the training and validation sets and test it on the test set. \textbf{\textit{DOTA}} series~\cite{dotav1.0} are oriented object detection datasets for remote sensing images. They include DOTA-v1.0~\cite{dotav1.0}/ v1.5 / v2.0~\cite{dota-v2.0}, which differ in the number of images, instances, and categories.
\TableExpSetting
The images range in size from 800$\times$800 to 4,000$\times$4,000 pixels and cover various scenes and objects. The categories of \textbf{\textit{DOTA-v1.0}} are 15. It contains 2,806 images and 188,282 instances. The categories of \textbf{\textit{DOTA-v1.5}} are 16. It uses the same images as DOTA-v1.0 but adds more small instances, resulting in 403,318 instances in total. The categories of \textbf{\textit{DOTA-v2.0}} are 18. It contains 11,268 images and 1,793,658 instances. We train our model on the training and validation sets of these dataset and test it on the test set. We submit test results to the official evaluation server of DOTA to obtain the detection performance. \textbf{\textit{HRSC2016}}~\cite{hrsc2016} is a challenging dataset for ship detection in remote sensing images, including 1,061 images. The dataset is divided into two sets: training and testing with 617 and 444 images respectively. We evaluate our model on the test set using two metrics, the PASCAL VOC07 and VOC12. \textbf{\textit{ICDAR2015}}~\cite{icdar2015} is utilized for text detection and comprises 1,000 training images along with 500 test images.

\DIORRResults
\DOTAOneResults

\subsection{Implementation Details and Evaluation Metrics}
\textbf{\textit{1) Implementation Details:}}
We conduct all the experiments on two NVIDIA RTX 2080ti with a batch of 4 (2 images per GPU). Models are constructed based on MMRotate~\cite{mmrotate} with Pytorch. The ResNet~\cite{resnet}, Swin~\cite{swin}, and LSK~\cite{lsknet} are used as backbone pre-trained on the ImageNet~\cite{imagenet}. We optimize models with AdamW optimizer~\cite{adamw} with learning rate $5\times e^{-5}$. The weights of losses are 5.0, 2.0, and 2.0 for Rotated IoU loss~\cite{rotated_iou_loss}, Focal loss~\cite{focal_loss}, and L1 loss respectively. Data augmentation strategies contain only random flips. Details of experiments are displayed in Table~\ref{expsetting} and Algo.~\ref{Training_prodedure}.

For experiments on the DOTA-v1.0/1.5/2.0, images are cropped into patches of $1024\times1024$ with overlaps of 200 and trained for 12 epochs. At epochs 8 and 11, the learning rate is divided by 10. In addition, for the multi-scale training in DOTA-v1.0, images are first resized at three scales (0.5, 1.0, and 1.5) and then cropped following single-scale training. For experiments on DIOR-R, images are trained for 12 epochs with the original fixed size of 800$\times$800. For experiments on HRSC2016, we scale images to a range of (512, 800) and train them for 24 epochs. Images on ICDAR2015 are trained for 24 epochs with a fixed size of 800$\times$800.

\textbf{\textit{2) Evaluation Metrics:}} $\mathrm{AP}_{50}$, $\mathrm{AP}_{75}$, and $\mathrm{AP}_{50:95}$ measure the accuracy of methods. We also analyze precision, recall, F-measure, PASCAL VOC 07 and 12 metrics for different methods. FPS is a metric for assessing inference speed. Params and FLOPs are used to count the parameters and complexity respectively of the model. Epochs are used to measure model training time.

\subsection{Comparisons With State-of-the-Arts}

\prcurveiouzeropointfivedior
\TableDOTAVonefiveResult
\TableResultICDAR

\prcurveiouzeropointsevenfivedior

\HRSCResults
\TableDOTAVtwoResult

\textbf{\textit{1) Results on DIOR-R:}} We compare OrientedFormer with modern CNN-based and transformer-based detectors. The detailed results of every category on the DIOR-R~\cite{dior} are reported in Table~\ref{DIOR-R-result} and Fig. \ref{pr_curve_iou0.5_dior}, \ref{pr_curve_iou0.75_dior}. Results of compared methods are from their papers. Our method achieves 65.07\% AP$_{50}$ with LSK-T, 67.28\% AP$_{50}$ with ResNet50, and 68.84\% AP$_{50}$ with Swin-T, outperforming all comparison CNN-based one-stage and two-stage detectors and transformer-based detectors.
%Specifically, OrientedFormer outperforms DCFL by 0.48\% (67.28\% VS 66.80\%), DODet by 2.18\% (67.28\% VS 64.41\%), and ARS-DETR by 1.16\% (67.28\% VS 66.12\%) with ResNet50, which is a large margin.
% In particular, we find that our method performs better in those categories with large aspect ratios, e.g. airport (APO), ship (SH), and vehicle (VE). Taking the ship for an example, our method improves DCFL, DODet, and ARS-DETR with the mAP gains of 3.46\%, 3.25\%, and 4.11\% respectively. It proves our method can predict accurately, even for objects with extreme aspect ratios.

\textbf{\textit{2) Results on DOTA-v1.0:}} We report results on DOTA-v1.0 in Table~\ref{DOTA-1.0-result}, compared with current CNN-based and transformer-based detectors. Results of compared methods are from their papers. In terms of accuracy measured by AP$_{50}$, OrientedFormer achieves 75.37\% AP$_{50}$ with ResNet50, 75.88\% AP$_{50}$ with Swin-T, and 75.92\% AP$_{50}$ with ResNet101 under single-scale data. Additionally, it achieves 79.06\% AP$_{50}$ with ResNet50 under multi-scale data.
% It achieves the best result among CNN-based one-stage and transformer-based methods. Specifically, with single-scale data, OrientedFormer outperforms SASM by 0.45\% (75.37\% VS 74.92\%) and RRoIFormer by 1.21\% (75.37\% VS 74.16\%). Compared with CNN-based two-stage methods, our method outperforms most of them, except for the Oriented R-CNN and SCRDet++. We argue that reasons are the characteristic of datasets and complex architecture. Different from these methods, our method does not require hand-craft components.

\textbf{\textit{3) Results on DOTA-v1.5:}} Table~\ref{DOTA-1.5-result} shows a comparison of our method with other modern detectors, using results from their papers. Using the ResNet50 backbone, our method achieves 67.06\% AP$_{50}$ with single-scale data. DOTA-v1.5 contains many small object instances, e.g., small vehicle (SV), ship (SH), and swimming pool (SP). For these instances, our method performs better.
% Taking the small vehicle as an example, our method improves ReDet with the mAP gains of 11.67\%. The results on the ship (85.33\%) and swimming pool (72.08) are also superior to other methods. It proves that our method is very effective for detecting tiny oriented objects.

\textbf{\textit{4) Results on ICDAR2015:}} We conduct experiments using the ICDAR2015~\cite{icdar2015}, shown in Table~\ref{ResultICDAR2015}. We reimplement compared methods using MMRotate, with the same settings as our methods. Our OrientedFormer achieves a precision of 85.3\%, a recall of 74.2\%, and an F-measure of 79.4\%.

\TableAblationStudiesIntegrated
\PositionalEncoding

\EachProposedModuleOnDIOR

\textbf{\textit{5) Results on HRSC2016:}} The HRSC2016 contains only ships. Table~\ref{hrsc2016_results} shows the results of our method and other object detectors from their papers. Our OrientedFormer achieves 90.17\% and 96.48\% AP$_{50}$ with ResNet50 under VOC07 and VOC12 metrics, respectively, which are competitive with modern detectors.
%Because only one category and less number of images in HRSC2016, modern detectors have all achieved high accuracy and similar results, e.g. around 90\% mAP under VOC07 metrics.

\textbf{\textit{6) Results on DOTA-v2.0:}}. As shown in Table~\ref{DOTA-2.0-result}, our proposed OrientedFormer is compared with CNN-based one-stage and two-stage detectors, using results reported in their papers. For a fair comparison, the backbones of all models are ResNet50. Our method achieves the state-of-the-art performance of 54.27\% AP$_{50}$ on the DOTA-v2.0 benchmark with single-scale data. Our method outperforms all compared CNN-based detectors.
%Overall, our method obtains 2.2\%, 2.2\%, and 3.91\% improvements in mAP over RRoIFormer, Oriented R-CNN, and DCFL. In particular, it can be found that our method performs better in detecting categories with small sizes, square-like shapes, and large aspect ratios. Our method achieves mAP of 56.54\% and 26.97\% on the vehicle (SV) and container crane (CC) with a small size. On the categories with square-like shapes, it achieves mAP of 54.58\% and 72.16\% on baseball diamond (BD) and storage tank (ST). On the categories with large aspect ratios, it achieves mAP of 78.53\% and 53.77\% on plane and harbor.

\subsection{Ablation Study}

% In this section, to demonstrate the effectiveness of our proposed method, we conduct a series of ablation studies on the DIOR-R\cite{dior}, with ResNet50 as the backbone and $1\times$ training scheme.

\textbf{\textit{1) Numbers of object queries:}} In this experiment, we evaluate the impact of the number of object queries, as illustrated in Table~\ref{ablations} (a). We noticed a significant improvement in performance as the number of object queries increased. With 100 object queries, the AP$_{50}$ is only 65.16\%, but when the number of object queries is increased to 300, it rises to 67.28\% (an improvement of 2.12\%). This suggests that a sufficient number of object queries can effectively cover the objects in images.
% However, when the number of object queries is further increased to 500, we observe a slight decrease in performance. We hypothesize that object queries are redundant.

\textbf{\textit{2) Number of sampling points in oriented cross-attention:}}
 %Following ~\cite{deformabledetr, adamixer}, our oriented cross-attention adopts numerous sampling points.
 In this ablation study, we use a different number of sampling points, shown in Table~\ref{ablations} (b). The reason why numerous sampling points are used is that features sampled by these sampling points are responsible for the classification and regression of objects, and spatial-aware attention mainly focuses on these features. As we increase the number of sampling points from 4 to 32, the AP$_{50}$ grows from 65.60\% to 67.28\%. It indicates that abundant features facilitate spatial-aware attention and the entire decoder.
 %As we continue to increase the number of sampling points, there is a decline in performance. We speculate that the redundant features are harmful to the attention.

\textbf{\textit{3) Number of attention heads in oriented cross-attention:}}
%Following RRoIFormer~\cite{rroiformer}, our oriented cross-attention adopts multiple heads.
In this experiment, we use different numbers of attention heads, shown in Table~\ref{ablations} (c). The reason why it is necessary to use multiple heads is that different heads establish different associations between queries and values. As we increase the number of heads from 8 to 64, the AP$_{50}$ grows from 66.33\% to 67.28\%. It indicates that heads in attention can provide multiple subspaces for representation, and extends the ability to focus on different parts of features.
%As we continue to increase the number of heads, there is a decline in performance. We analyze that redundant heads focus on redundant features.

\AttentionBias
\EachProposedModuleOnDOTA

\subsection{Comparisons of different Positional Encodings}
We perform ablation experiments with different PE, shown in Table~\ref{ablation_pe}. The learnable PE~\cite{detr}, sinusoidal absolute PE following Deformable DETR~\cite{deformabledetr}, and DAB DETR~\cite{dab_detr} are compared. They only encode sizes and positions of oriented boxes, but lack angles. The model achieves 66.85\% without any PE, due to a lack of information of sequence ordering. When the PE of Deformable DETR and DAB DETR, and the learnable one are used, the AP$_{50}$ is reduced. We argue that these PEs are mismatched for positional queries. Our Gaussian PE can bring an improvement in performance.

\subsection{Comparisons of different Self-Attention}
In Table~\ref{ablation_self_attn}, we compare Wasserstein self-attention with other modern self-attention. When vanilla self-attention~\cite{detr,deformabledetr} is applied, the model only achieves an AP$_{50}$ of 67.03\%. We replace Gaussian Wasserstein distance scores with Intersection over Foreground (IoF) and Intersection over Union (IoU) for comparison.
Compared with IoF and IoU methods, our proposed Wasserstein self-attention yields the best performance, achieving improvements of 0.71\% and 0.2\% respectively.

\convergence
\compareepochsmap
\TableFPS
\Tabblelevels
\TableSamplingOffsets

\subsection{Effects of proposed Individual Strategy}
In this study, we evaluate the effectiveness of each strategy proposed in our method, including Gaussian positional encoding (PE), Wasserstein self-attention, and oriented cross-attention, as shown in Table~\ref{ablation_indivisual_on_DIOR} and~\ref{ablation_indivisual_on_DOTA}. Incrementally incorporating each individual strategies, they improve performance.

\subsection{Convergence and Traning Epochs}
DETR~\cite{detr} suffers from slow convergence and long training times. To further investigate convergence, we compare OrientedFormer with other end-to-end models, depicted in Fig.~\ref{convergence_curve}. For fair comparisons, all methods are trained on 12 epochs on DIOR-R with 300 queries. OrientedForemr achieves an AP$_{50}$ of 67.3\% in just 12 epochs with ResNet50, surpassing Deformable DETR-O with CSL (31.2\%) and ARS-DETR (38.9\%).

We compare OrientedFormer with other end-to-end models for accuracy and training epochs, shown in Fig.~\ref{epochs_vs_map}. Training OrientedFormer for 12 epochs can outperform ARS-DETR and Deformable DETR-O, which require 36 epochs of training. Specifically, OrientedFormer achieves an AP$_{50}$ of 75.37\% with ResNet50 in 12 epochs, while ARS-DETR achieves only 74.16\% and Deformable DETR-O achieves only 69.48\% in 36 epochs.

\comparewithothermethods
\differentsamplingpoints
\querycenterpoints

\subsection{Comparison of Speed, Parameters, FLOPs and Accuracy}
To further explore the performance of OrientedFormer, we conducted an experiment comparing its FPS, parameters, FLOPs, and AP$_{50}$ with other methods on DOTA-v1.0, shown in Table~\ref{FPS_Params}. OrientedFormer outpaces CNN-based two-stage methods and other end-to-end methods in speed but slightly trails behind most one-stage methods. Additionally, OrientedFormer has a slightly larger number of parameters compared to other end-to-end methods.

\subsection{Comparison of Different Feature Layers of Backbone}
Backbones extract features from images and play an important role in oriented object detection. We conducted an experiment with different feature layers of backbone ResNet50, shown in Table~\ref{ablation_levels}. As the number of backbone network feature layers increases, the AP, parameters, and FLOPs of the model gradually increase. 4 layers of features are selected because multi-scale features can capture rich information from the images.

\visualsamplingpoint
\visualresults
\suboptimalresults

\vspace{-3mm}
\subsection{Comparison of Different Sampling Methods}
Learnable sampling points $P$ are rotated by angles of positional queries for alignment in our oriented cross-attention. We compare our methods with different sampling strategies, shown in Table~\ref{SamplingOffsets}. The Fixed offset method utilizes fixed sampling points around the center points of positional queries. The Deformable offsets method does not rotate sampling points, which is the main difference between our methods. The Random offsets method employs random sampling points around the center points of positional queries. We visualize sampling points of different methods in Fig.~\ref{differentsamplingpoints}. Our oriented cross-attention can more effectively align features and thus focus on more accurate object features.

\subsection{Visualization}
\textbf{\textit{1) Comparison with other methods:}} We compare our method with others on large-scale objects, densely packed objects, images with complex backgrounds, and images under low lighting conditions, shown in Fig.~\ref{compare_with_other_method}. Other methods struggle with accurately detecting large objects, often miss densely packed objects, mistake background noise for objects, and fail to perform well under poor environmental conditions.

\textbf{\textit{2) Learnable positional queries:}} Positional queries are used for locations of objects. We visualize the center points of learned positional queries shown in Fig.~\ref{querycenterpoints}. Default 300 positional queries are used in the experiments. Positional queries are positioned at the center of objects and suspected objects. It demonstrates the utility of positional queries for representing object positions.

\textbf{\textit{3) Learnable sampling points:}} For a better understanding learned oriented cross-attention, we visualize centers of positional queries and sampling points in the decoder, as shown in Fig.~\ref{fig_visual_sampling_point}. For readability, all sampling points are scaled to original images. Features sampled from sampling points play the role of values in cross-attention. Sampling points have aligned with oriented boxes. Most of the sampling points are distributed within oriented boxes and others out of boxes.

\textbf{\textit{4) Detection results:}} We visualize detection results on different datasets. Fig.~\ref{fig_visual_results} shows the detection results on DOTA, DIOR-R, and HRSC2016. Oriented boxes locate objects accurately in images. It is worth noting that the DOTA dataset contains numerous images depicting extreme weather and poor lighting conditions. However, our method remains effective for detection under these circumstances.

\textbf{\textit{5) Suboptimal results:}} We show some suboptimal results in Fig.~\ref{suboptimal results}. There are a large number of objects with small size and huge aspect ratios in oriented remote sensing images. In addition, some foreground objects are similar to the background. These challenges warrant further investigation in future research.

\section{Conclusions}
In this paper, we propose an end-to-end transformer-based detector OrientedFormer for oriented object detection in remote sensing images. The proposed OrientedFormer comprises the Gaussian positional encoding, Wasserstein self-attention, and oriented cross-attention. These dedicated components work together to accurately classify and localize objects with multiple orientations in remote sensing images. First, Gaussian positional encoding is introduced to encode not only the position and size but also the angles of oriented boxes. Second, Wasserstein self-attention is proposed to incorporate geometric relations between content queries into the self-attention mechanism. Finally, oriented cross-attention is designed to address misalignment issues by rotating sampling points according to object angles. Extensive experiments on six datasets demonstrate the effectiveness of our methods. We validate that transformer-based detectors can be competitive with CNN-based one-stage and two-stage detectors. In comparison to previous end-to-end detectors, the OrientedFormer achieves a performance increase of 1.16 and 1.21 AP$_{50}$ on DIOR-R and DOTA-v1.0 respectively, while also reducing training epochs from 3$\times$ to 1$\times$.

\textit{Limitations}. We defer the task of further reducing the number of parameters and enhancing the inference speed of the model to future research endeavors. In oriented object detection, the rotation-equivariant network is sensitive to the orientations of objects. In the future, we hope to construct an end-to-end transformer-based rotation-equivariant-oriented object detector.

\bibliographystyle{IEEEtran}
\bibliography{reference}

@INPROCEEDINGS{9925528,
  author={Biswas, Debojyoti and Rahman, M M Mahabubur and Zong, Ziliang and Tešić, Jelena},
  booktitle={2022 IEEE International Conference on Networking, Architecture and Storage (NAS)},
  title={Improving the Energy Efficiency of Real-time DNN Object Detection via Compression, Transfer Learning, and Scale Prediction},
  year={2022},
  volume={},
  number={},
  pages={1-8},
  doi={10.1109/NAS55553.2022.9925528}}

@INPROCEEDINGS{10008383,
  author={Biswas, Debojyoti and Tešić, Jelena},
  booktitle={2022 14th International Conference on Computational Intelligence and Communication Networks (CICN)},
  title={Small Object Difficulty (SOD) Modeling for Objects Detection in Satellite Images},
  year={2022},
  volume={},
  number={},
  pages={125-130},
  doi={10.1109/CICN56167.2022.10008383}}

@ARTICLE{10380690,
  author={Biswas, Debojyoti and Tešić, Jelena},
  journal={IEEE Journal of Selected Topics in Applied Earth Observations and Remote Sensing},
  title={Unsupervised Domain Adaptation With Debiased Contrastive Learning and Support-Set Guided Pseudolabeling for Remote Sensing Images},
  year={2024},
  volume={17},
  number={},
  pages={3197-3210},
  doi={10.1109/JSTARS.2024.3349541}}

@inproceedings{dcfl,
  title={Dynamic Coarse-to-Fine Learning for Oriented Tiny Object Detection},
  author={Xu, Chang and Ding, Jian and Wang, Jinwang and Yang, Wen and Yu, Huai and Yu, Lei and Xia, Gui-Song},
  booktitle={Proceedings of the IEEE/CVF Conference on Computer Vision and Pattern Recognition},
  pages={7318--7328},
  year={2023}
}

@inproceedings{psc,
  title={Phase-shifting coder: Predicting accurate orientation in oriented object detection},
  author={Yu, Yi and Da, Feipeng},
  booktitle={Proceedings of the IEEE/CVF Conference on Computer Vision and Pattern Recognition},
  pages={13354--13363},
  year={2023}
}

@inproceedings{orientedrcnn,
  title={Oriented R-CNN for object detection},
  author={Xie, Xingxing and Cheng, Gong and Wang, Jiabao and Yao, Xiwen and Han, Junwei},
  booktitle={Proceedings of the IEEE/CVF international conference on computer vision},
  pages={3520--3529},
  year={2021}
}

@inproceedings{redet,
  title={Redet: A rotation-equivariant detector for aerial object detection},
  author={Han, Jiaming and Ding, Jian and Xue, Nan and Xia, Gui-Song},
  booktitle={Proceedings of the IEEE/CVF Conference on Computer Vision and Pattern Recognition},
  pages={2786--2795},
  year={2021}
}

@ARTICLE{arsdetr,
  author={Zeng, Ying and Chen, Yushi and Yang, Xue and Li, Qingyun and Yan, Junchi},
  journal={IEEE Transactions on Geoscience and Remote Sensing},
  title={ARS-DETR: Aspect Ratio-Sensitive Detection Transformer for Aerial Oriented Object Detection},
  year={2024},
  volume={62},
  number={},
  pages={1-15},
  doi={10.1109/TGRS.2024.3364713}}

@inproceedings{detr,
  title={End-to-end object detection with transformers},
  author={Carion, Nicolas and Massa, Francisco and Synnaeve, Gabriel and Usunier, Nicolas and Kirillov, Alexander and Zagoruyko, Sergey},
  booktitle={European conference on computer vision},
  pages={213--229},
  year={2020},
  organization={Springer}
}

@inproceedings{resnet,
  title={Deep residual learning for image recognition},
  author={He, Kaiming and Zhang, Xiangyu and Ren, Shaoqing and Sun, Jian},
  booktitle={Proceedings of the IEEE conference on computer vision and pattern recognition},
  pages={770--778},
  year={2016}
}

@article{psd-sq,
  title={PSD-SQ: Point Set Decoding Based on Semantic Query for Object Detection in Remote Sensing Images},
  author={Feng, Shiyang and Wang, Bin},
  journal={IEEE Transactions on Geoscience and Remote Sensing},
  year={2024},
  publisher={IEEE}
}

@inproceedings{d2qdetr,
  title={D2Q-DETR: Decoupling and Dynamic Queries for Oriented Object Detection with Transformers},
  author={Zhou, Qiang and Yu, Chaohui and Wang, Zhibin and Wang, Fan},
  booktitle={ICASSP 2023-2023 IEEE International Conference on Acoustics, Speech and Signal Processing (ICASSP)},
  pages={1--5},
  year={2023},
  organization={IEEE}
}

@article{emo2detr,
  title={EMO2-DETR: Efficient-Matching Oriented Object Detection with Transformers},
  author={Hu, Zibo and Gao, Kun and Zhang, Xiaodian and Wang, Junwei and Wang, Hong and Yang, Zhijia and Li, Chenrui and Li, Wei},
  journal={IEEE Transactions on Geoscience and Remote Sensing},
  year={2023},
  publisher={IEEE}
}

@inproceedings{deformabledetr,
  title={Deformable DETR: Deformable Transformers for End-to-End Object Detection},
  author={Zhu, Xizhou and Su, Weijie and Lu, Lewei and Li, Bin and Wang, Xiaogang and Dai, Jifeng},
  booktitle={International Conference on Learning Representations},
  year={2020}
}

@article{s2anet,
  title={Align deep features for oriented object detection},
  author={Han, Jiaming and Ding, Jian and Li, Jie and Xia, Gui-Song},
  journal={IEEE Transactions on Geoscience and Remote Sensing},
  volume={60},
  pages={1--11},
  year={2021},
  publisher={IEEE}
}

@inproceedings{dotav1.0,
  title={DOTA: A large-scale dataset for object detection in aerial images},
  author={Xia, Gui-Song and Bai, Xiang and Ding, Jian and Zhu, Zhen and Belongie, Serge and Luo, Jiebo and Datcu, Mihai and Pelillo, Marcello and Zhang, Liangpei},
  booktitle={Proceedings of the IEEE conference on computer vision and pattern recognition},
  pages={3974--3983},
  year={2018}
}

@inproceedings{ddq,
  title={Dense Distinct Query for End-to-End Object Detection},
  author={Zhang, Shilong and Wang, Xinjiang and Wang, Jiaqi and Pang, Jiangmiao and Lyu, Chengqi and Zhang, Wenwei and Luo, Ping and Chen, Kai},
  booktitle={Proceedings of the IEEE/CVF Conference on Computer Vision and Pattern Recognition},
  pages={7329--7338},
  year={2023}
}

@inproceedings{roi_transformer,
  title={Learning RoI transformer for oriented object detection in aerial images},
  author={Ding, Jian and Xue, Nan and Long, Yang and Xia, Gui-Song and Lu, Qikai},
  booktitle={Proceedings of the IEEE/CVF Conference on Computer Vision and Pattern Recognition},
  pages={2849--2858},
  year={2019}
}

@inproceedings{smca,
  title={Fast convergence of detr with spatially modulated co-attention},
  author={Gao, Peng and Zheng, Minghang and Wang, Xiaogang and Dai, Jifeng and Li, Hongsheng},
  booktitle={Proceedings of the IEEE/CVF international conference on computer vision},
  pages={3621--3630},
  year={2021}
}

@inproceedings{anchor_detr,
  title={Anchor detr: Query design for transformer-based detector},
  author={Wang, Yingming and Zhang, Xiangyu and Yang, Tong and Sun, Jian},
  booktitle={Proceedings of the AAAI conference on artificial intelligence},
  volume={36},
  number={3},
  pages={2567--2575},
  year={2022}
}

@inproceedings{dynamic_detr,
  title={Dynamic detr: End-to-end object detection with dynamic attention},
  author={Dai, Xiyang and Chen, Yinpeng and Yang, Jianwei and Zhang, Pengchuan and Yuan, Lu and Zhang, Lei},
  booktitle={Proceedings of the IEEE/CVF International Conference on Computer Vision},
  pages={2988--2997},
  year={2021}
}

@inproceedings{dynamic_conv,
  title={Dynamic convolution: Attention over convolution kernels},
  author={Chen, Yinpeng and Dai, Xiyang and Liu, Mengchen and Chen, Dongdong and Yuan, Lu and Liu, Zicheng},
  booktitle={Proceedings of the IEEE/CVF conference on computer vision and pattern recognition},
  pages={11030--11039},
  year={2020}
}

@article{faster_rcnn,
  title={Faster r-cnn: Towards real-time object detection with region proposal networks},
  author={Ren, Shaoqing and He, Kaiming and Girshick, Ross and Sun, Jian},
  journal={Advances in neural information processing systems},
  volume={28},
  year={2015}
}

@inproceedings{atss,
  title={Bridging the gap between anchor-based and anchor-free detection via adaptive training sample selection},
  author={Zhang, Shifeng and Chi, Cheng and Yao, Yongqiang and Lei, Zhen and Li, Stan Z},
  booktitle={Proceedings of the IEEE/CVF conference on computer vision and pattern recognition},
  pages={9759--9768},
  year={2020}
}

@inproceedings{focal_loss,
  title={Focal loss for dense object detection},
  author={Lin, Tsung-Yi and Goyal, Priya and Girshick, Ross and He, Kaiming and Doll{\'a}r, Piotr},
  booktitle={Proceedings of the IEEE international conference on computer vision},
  pages={2980--2988},
  year={2017}
}

@article{dior,
  title={Anchor-free oriented proposal generator for object detection},
  author={Cheng, Gong and Wang, Jiabao and Li, Ke and Xie, Xingxing and Lang, Chunbo and Yao, Yanqing and Han, Junwei},
  journal={IEEE Transactions on Geoscience and Remote Sensing},
  volume={60},
  pages={1--11},
  year={2022},
  publisher={IEEE}
}

@inproceedings{imagenet,
  title={Imagenet: A large-scale hierarchical image database},
  author={Deng, Jia and Dong, Wei and Socher, Richard and Li, Li-Jia and Li, Kai and Fei-Fei, Li},
  booktitle={2009 IEEE conference on computer vision and pattern recognition},
  pages={248--255},
  year={2009},
  organization={Ieee}
}

@inproceedings{adamw,
  title={Decoupled Weight Decay Regularization},
  author={Loshchilov, Ilya and Hutter, Frank},
  booktitle={International Conference on Learning Representations},
  year={2018}
}

@inproceedings{r3det,
  title={R3det: Refined single-stage detector with feature refinement for rotating object},
  author={Yang, Xue and Yan, Junchi and Feng, Ziming and He, Tao},
  booktitle={Proceedings of the AAAI conference on artificial intelligence},
  volume={35},
  number={4},
  pages={3163--3171},
  year={2021}
}

@inproceedings{cfa,
  title={Beyond bounding-box: Convex-hull feature adaptation for oriented and densely packed object detection},
  author={Guo, Zonghao and Liu, Chang and Zhang, Xiaosong and Jiao, Jianbin and Ji, Xiangyang and Ye, Qixiang},
  booktitle={Proceedings of the IEEE/CVF conference on Computer Vision and Pattern Recognition},
  pages={8792--8801},
  year={2021}
}

@inproceedings{sasm,
  title={Shape-adaptive selection and measurement for oriented object detection},
  author={Hou, Liping and Lu, Ke and Xue, Jian and Li, Yuqiu},
  booktitle={Proceedings of the AAAI Conference on Artificial Intelligence},
  volume={36},
  number={1},
  pages={923--932},
  year={2022}
}

@inproceedings{h2rbox,
  title={H2RBox: Horizontal Box Annotation is All You Need for Oriented Object Detection},
  author={Yang, Xue and Zhang, Gefan and Li, Wentong and Zhou, Yue and Wang, Xuehui and Yan, Junchi},
  booktitle={The Eleventh International Conference on Learning Representations},
  year={2022}
}

@article{gliding_vertex,
  title={Gliding vertex on the horizontal bounding box for multi-oriented object detection},
  author={Xu, Yongchao and Fu, Mingtao and Wang, Qimeng and Wang, Yukang and Chen, Kai and Xia, Gui-Song and Bai, Xiang},
  journal={IEEE transactions on pattern analysis and machine intelligence},
  volume={43},
  number={4},
  pages={1452--1459},
  year={2020},
  publisher={IEEE}
}

@inproceedings{oriented_reppoints,
  title={Oriented reppoints for aerial object detection},
  author={Li, Wentong and Chen, Yijie and Hu, Kaixuan and Zhu, Jianke},
  booktitle={Proceedings of the IEEE/CVF conference on computer vision and pattern recognition},
  pages={1829--1838},
  year={2022}
}

@article{QPDet,
  title={On improving bounding box representations for oriented object detection},
  author={Yao, Yanqing and Cheng, Gong and Wang, Guangxing and Li, Shengyang and Zhou, Peicheng and Xie, Xingxing and Han, Junwei},
  journal={IEEE Transactions on Geoscience and Remote Sensing},
  volume={61},
  pages={1--11},
  year={2022},
  publisher={IEEE}
}

@inproceedings{rotated_iou_loss,
  title={Iou loss for 2d/3d object detection},
  author={Zhou, Dingfu and Fang, Jin and Song, Xibin and Guan, Chenye and Yin, Junbo and Dai, Yuchao and Yang, Ruigang},
  booktitle={2019 international conference on 3D vision (3DV)},
  pages={85--94},
  year={2019},
  organization={IEEE}
}

@article{dota-v2.0,
  title={Object detection in aerial images: A large-scale benchmark and challenges},
  author={Ding, Jian and Xue, Nan and Xia, Gui-Song and Bai, Xiang and Yang, Wen and Yang, Michael Ying and Belongie, Serge and Luo, Jiebo and Datcu, Mihai and Pelillo, Marcello and others},
  journal={IEEE transactions on pattern analysis and machine intelligence},
  volume={44},
  number={11},
  pages={7778--7796},
  year={2021},
  publisher={IEEE}
}

@inproceedings{csl,
  title={Arbitrary-oriented object detection with circular smooth label},
  author={Yang, Xue and Yan, Junchi},
  booktitle={Computer Vision--ECCV 2020: 16th European Conference, Glasgow, UK, August 23--28, 2020, Proceedings, Part VIII 16},
  pages={677--694},
  year={2020},
  organization={Springer}
}

@inproceedings{mask_rcnn,
  title={Mask r-cnn},
  author={He, Kaiming and Gkioxari, Georgia and Doll{\'a}r, Piotr and Girshick, Ross},
  booktitle={Proceedings of the IEEE international conference on computer vision},
  pages={2961--2969},
  year={2017}
}

@inproceedings{htc,
  title={Hybrid task cascade for instance segmentation},
  author={Chen, Kai and Pang, Jiangmiao and Wang, Jiaqi and Xiong, Yu and Li, Xiaoxiao and Sun, Shuyang and Feng, Wansen and Liu, Ziwei and Shi, Jianping and Ouyang, Wanli and others},
  booktitle={Proceedings of the IEEE/CVF conference on computer vision and pattern recognition},
  pages={4974--4983},
  year={2019}
}

@inproceedings{mmrotate,
  title={Mmrotate: A rotated object detection benchmark using pytorch},
  author={Zhou, Yue and Yang, Xue and Zhang, Gefan and Wang, Jiabao and Liu, Yanyi and Hou, Liping and Jiang, Xue and Liu, Xingzhao and Yan, Junchi and Lyu, Chengqi and others},
  booktitle={Proceedings of the 30th ACM International Conference on Multimedia},
  pages={7331--7334},
  year={2022}
}

@article{attention_is_all_you_need,
  title={Attention is all you need},
  author={Vaswani, Ashish and Shazeer, Noam and Parmar, Niki and Uszkoreit, Jakob and Jones, Llion and Gomez, Aidan N and Kaiser, {\L}ukasz and Polosukhin, Illia},
  journal={Advances in neural information processing systems},
  volume={30},
  year={2017}
}

@inproceedings{adamixer,
  title={Adamixer: A fast-converging query-based object detector},
  author={Gao, Ziteng and Wang, Limin and Han, Bing and Guo, Sheng},
  booktitle={Proceedings of the IEEE/CVF Conference on Computer Vision and Pattern Recognition},
  pages={5364--5373},
  year={2022}
}

@inproceedings{mip_nerf,
  title={Mip-nerf: A multiscale representation for anti-aliasing neural radiance fields},
  author={Barron, Jonathan T and Mildenhall, Ben and Tancik, Matthew and Hedman, Peter and Martin-Brualla, Ricardo and Srinivasan, Pratul P},
  booktitle={Proceedings of the IEEE/CVF International Conference on Computer Vision},
  pages={5855--5864},
  year={2021}
}

@inproceedings{gwd,
  title={Rethinking rotated object detection with gaussian wasserstein distance loss},
  author={Yang, Xue and Yan, Junchi and Ming, Qi and Wang, Wentao and Zhang, Xiaopeng and Tian, Qi},
  booktitle={International conference on machine learning},
  pages={11830--11841},
  year={2021},
  organization={PMLR}
}

@inproceedings{
  dab_detr,
  title={{DAB}-{DETR}: Dynamic Anchor Boxes are Better Queries for {DETR}},
  author={Shilong Liu and Feng Li and Hao Zhang and Xiao Yang and Xianbiao Qi and Hang Su and Jun Zhu and Lei Zhang},
  booktitle={International Conference on Learning Representations},
  year={2022}
}

@article{hungarian_matching,
  title={The Hungarian method for the assignment problem},
  author={Kuhn, Harold W},
  journal={Naval research logistics quarterly},
  volume={2},
  number={1-2},
  pages={83--97},
  year={1955},
  publisher={Wiley Online Library}
}

@article{hrsc2016,
  title={Ship rotated bounding box space for ship extraction from high-resolution optical satellite images with complex backgrounds},
  author={Liu, Zikun and Wang, Hongzhen and Weng, Lubin and Yang, Yiping},
  journal={IEEE geoscience and remote sensing letters},
  volume={13},
  number={8},
  pages={1074--1078},
  year={2016},
  publisher={IEEE}
}

@article{ao2detr,
  title={Ao2-detr: Arbitrary-oriented object detection transformer},
  author={Dai, Linhui and Liu, Hong and Tang, Hao and Wu, Zhiwei and Song, Pinhao},
  journal={IEEE Transactions on Circuits and Systems for Video Technology},
  year={2022},
  publisher={IEEE}
}

@inproceedings{h2rboxv2,
  title={H2RBox-v2: Incorporating Symmetry for Boosting Horizontal Box Supervised Oriented Object Detection},
  author={Yu, Yi and Yang, Xue and Li, Qingyun and Zhou, Yue and Da, Feipeng and Yan, Junchi},
  booktitle={Thirty-seventh Conference on Neural Information Processing Systems},
  year={2023}
}

@inproceedings{oskdet,
  title={OSKDet: Orientation-sensitive keypoint localization for rotated object detection},
  author={Lu, Dongchen and Li, Dongmei and Li, Yali and Wang, Shengjin},
  booktitle={Proceedings of the IEEE/CVF Conference on Computer Vision and Pattern Recognition},
  pages={1182--1192},
  year={2022}
}

@inproceedings{piou,
  title={Piou loss: Towards accurate oriented object detection in complex environments},
  author={Chen, Zhiming and Chen, Kean and Lin, Weiyao and See, John and Yu, Hui and Ke, Yan and Yang, Cong},
  booktitle={Computer Vision--ECCV 2020: 16th European Conference, Glasgow, UK, August 23--28, 2020, Proceedings, Part V 16},
  pages={195--211},
  year={2020},
  organization={Springer}
}

@article{centermap,
  title={Learning center probability map for detecting objects in aerial images},
  author={Wang, Jinwang and Yang, Wen and Li, Heng-Chao and Zhang, Haijian and Xia, Gui-Song},
  journal={IEEE Transactions on Geoscience and Remote Sensing},
  volume={59},
  number={5},
  pages={4307--4323},
  year={2020},
  publisher={IEEE}
}

@InProceedings{lsknet,
    author    = {Li, Yuxuan and Hou, Qibin and Zheng, Zhaohui and Cheng, Ming-Ming and Yang, Jian and Li, Xiang},
    title     = {Large Selective Kernel Network for Remote Sensing Object Detection},
    booktitle = {Proceedings of the IEEE/CVF International Conference on Computer Vision (ICCV)},
    month     = {October},
    year      = {2023},
    pages     = {16794-16805}
}

@ARTICLE{dodet,
  author={Cheng, Gong and Yao, Yanqing and Li, Shengyang and Li, Ke and Xie, Xingxing and Wang, Jiabao and Yao, Xiwen and Han, Junwei},
  journal={IEEE Transactions on Geoscience and Remote Sensing},
  title={Dual-Aligned Oriented Detector},
  year={2022},
  volume={60},
  number={},
  pages={1-11},
  doi={10.1109/TGRS.2022.3149780}}

@ARTICLE {SCRDetplusplus,
author = {X. Yang and J. Yan and W. Liao and X. Yang and J. Tang and T. He},
journal = {IEEE Transactions on Pattern Analysis and Machine Intelligence},
title = {SCRDet++: Detecting Small, Cluttered and Rotated Objects via Instance-Level Feature Denoising and Rotation Loss Smoothing},
year = {2023},
volume = {45},
number = {02},
issn = {1939-3539},
pages = {2384-2399},
doi = {10.1109/TPAMI.2022.3166956},
publisher = {IEEE Computer Society},
address = {Los Alamitos, CA, USA},
month = {feb}
}

@ARTICLE{dfdet,
  author={Xie, Xingxing and Cheng, Gong and Rao, Chaofan and Lang, Chunbo and Han, Junwei},
  journal={IEEE Transactions on Geoscience and Remote Sensing},
  title={Oriented Object Detection via Contextual Dependence Mining and Penalty-Incentive Allocation},
  year={2024},
  volume={62},
  number={},
  pages={1-10},
  doi={10.1109/TGRS.2024.3385985}}

@ARTICLE{dhrec,
  author={Nie, Guangtao and Huang, Hua},
  journal={IEEE Transactions on Pattern Analysis and Machine Intelligence},
  title={Multi-Oriented Object Detection in Aerial Images With Double Horizontal Rectangles},
  year={2023},
  volume={45},
  number={4},
  pages={4932-4944},
  doi={10.1109/TPAMI.2022.3191753}}

@inproceedings{icdar2015,
  title={ICDAR 2015 competition on robust reading},
  author={Karatzas, Dimosthenis and Gomez-Bigorda, Lluis and Nicolaou, Anguelos and Ghosh, Suman and Bagdanov, Andrew and Iwamura, Masakazu and Matas, Jiri and Neumann, Lukas and Chandrasekhar, Vijay Ramaseshan and Lu, Shijian and others},
  booktitle={2015 13th international conference on document analysis and recognition (ICDAR)},
  pages={1156--1160},
  year={2015},
  organization={IEEE}
}

@inproceedings{swin,
  title={Swin transformer: Hierarchical vision transformer using shifted windows},
  author={Liu, Ze and Lin, Yutong and Cao, Yue and Hu, Han and Wei, Yixuan and Zhang, Zheng and Lin, Stephen and Guo, Baining},
  booktitle={Proceedings of the IEEE/CVF international conference on computer vision},
  pages={10012--10022},
  year={2021}
}

@ARTICLE{cledet,
  author={Zhou, Yong and Chen, Silin and Zhao, Jiaqi and Yao, Rui and Xue, Yong and Saddik, Abdulmotaleb El},
  journal={IEEE Transactions on Geoscience and Remote Sensing}, 
  title={CLT-Det: Correlation Learning Based on Transformer for Detecting Dense Objects in Remote Sensing Images}, 
  year={2022},
  volume={60},
  number={},
  pages={1-15},
  doi={10.1109/TGRS.2022.3204770}}

\end{document}